\documentclass{article} %
\usepackage{iclr2026_conference,times}

\usepackage{amsmath,amsfonts,bm}

\def\eqref#1{equation~\ref{#1}}

\def\1{\bm{1}}

\DeclareMathAlphabet{\mathsfit}{\encodingdefault}{\sfdefault}{m}{sl}
\SetMathAlphabet{\mathsfit}{bold}{\encodingdefault}{\sfdefault}{bx}{n}

\usepackage{hyperref}
\usepackage{url}
\usepackage{pifont}%
\usepackage{longtable}
\usepackage{colortbl}
\usepackage{xspace}
\usepackage{titlesec}
\usepackage{arydshln}

\usepackage{array}
\usepackage[most]{tcolorbox}
\usepackage{lipsum}
\usepackage{subcaption} %

\usepackage{longtable}  %
\usepackage{booktabs}  %
\usepackage{multirow}
\usepackage{tabularx}
\usepackage{cleveref}
\usepackage{tabularray}
\usepackage{wrapfig}
\usepackage{enumitem}

\usepackage[textsize=tiny]{todonotes}

\author{
Karmesh Yadav$^{1*}$
 \,\,\,
Yusuf Ali$^{1*}$
 \,\,\,
Gunshi Gupta$^{2}$
 \,\,\,
Yarin Gal$^{2}$
 \,\,\,
Zsolt Kira$^{1}$
 \\
$^1$Georgia Tech  \quad  $^2$University of Oxford
}

\definecolor{skyblue}{HTML}{7EC0EE}
\newcommand{\graycell}{\cellcolor{lightgray!10}}

\newcommand{\cmark}{\ding{51}}%
\newcommand{\xmark}{\ding{55}}%
\newcommand{\ourbench}{\textsc{FindingDory}\xspace}
\newcommand{\numtasks}{60\xspace}

\renewcommand{\arraystretch}{1.3}
\newcolumntype{L}{>{\columncolor{white}}l}

\title{\ourbench: A Benchmark to Evaluate\\Memory in Embodied Agents}

\titlespacing*{\section}{0pt}{0.9ex plus 0.1ex minus 0.3ex}{0.6ex plus 0.1ex minus 0.2ex}
\titlespacing*{\subsection}{0pt}{0.6ex plus 0.1ex minus 0.2ex}{0.4ex plus 0.1ex minus 0.2ex}

\iclrfinalcopy %
\begin{document}
\setlist[enumerate]{noitemsep,topsep=0pt}

\maketitle

\begin{abstract}

Vision-Language models (VLMs) have recently demonstrated impressive performance in planning and control tasks, driving interest in their application to robotics. Yet their deployment in embodied settings remains limited by the challenge of incorporating long-term experience, often spanning multiple days and represented by vast image collections. Current VLMs typically handle only a few hundred images at once, underscoring the need for more efficient mechanisms to manage long-term memory in embodied contexts.
To meaningfully evaluate these models for long-horizon control, a benchmark must target scenarios where memory is essential. Existing long-video QA benchmarks neglect embodied challenges like object manipulation and navigation, which require low-level skills and fine-grained reasoning over past interactions. Moreover, effective memory integration in embodied agents involves both recalling relevant historical information and executing actions based on that information, making it essential to study these aspects together.
In this work, we introduce \ourbench, a new benchmark for long-range embodied tasks in the Habitat simulator. \ourbench evaluates memory-centric capabilities across \numtasks tasks requiring sustained engagement and contextual awareness in an environment. The tasks can also be procedurally extended to longer and more challenging versions, enabling scalable evaluation of memory and reasoning. We further present baselines that integrate state-of-the-art closed-source and fine-tuned open-source VLMs with low-level navigation policies, assessing their performance on these memory-intensive tasks and highlighting key areas for improvement.\footnote{Website: \url{https://findingdory-benchmark.github.io/}}
\end{abstract}    
\section{Introduction}
\label{sec:intro}

\begin{figure}[t]
    \centering
    \includegraphics[width=\textwidth]{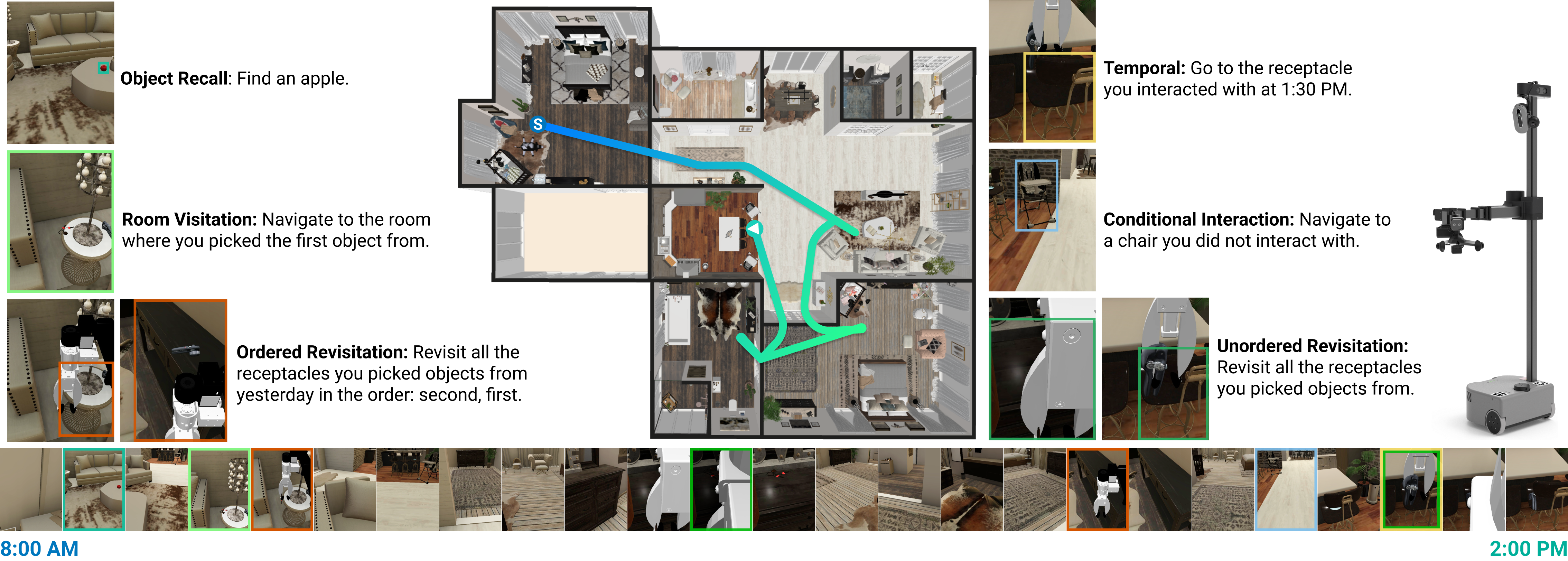}  
    \caption{\ourbench: We propose a benchmark for evaluating agents’ memory by constructing scenarios that require reasoning over previously collected experiences. In these scenarios, an agent receives logs of prior interactions in indoor environments and must complete navigation, pick, and place instructions. The figure shows a sample episode: first, a privileged agent with full environment access executes randomly sampled navigation and manipulation tasks. Then, the memory-evaluating agent is placed in the same environment and tasked with completing instructions that rely on the privileged agent’s logs to decide where to navigate or which object to retrieve.}
    \vspace{-18pt}
    \label{fig:main}
\end{figure}

Memory is a fundamental capability for intelligent agents that allows them to recall past experiences, adapt to dynamic environments, and make informed decisions over extended timescales. In humans and animals, it plays a crucial role in navigation, reasoning, and goal-directed behavior. As we strive to develop more capable embodied systems, equipping them with robust memory mechanisms is essential, particularly in settings where agents must process high-dimensional multimodal inputs to interact with their surroundings.
The ability to store and retrieve relevant information is the key to unlocking sophisticated behaviors, from household assistance to autonomous exploration.

Recent advances in Vision-Language Models (VLMs) have significantly improved high-level reasoning and planning for embodied tasks. These models leverage large-scale multimodal training to exhibit strong scene understanding and action-guided perception. However, most existing VLM applications focus on short-term or static tasks, such as image captioning or Visual Question Answering (VQA), which do not require sustained memory usage. Extending these models to long-horizon embodied control introduces new challenges: embodied agents must integrate observations over time, recall past experiences, and act accordingly within their environment.

While recent efforts have sought to extend long-context understanding in VLMs, primarily through video-based QA or long-document comprehension~\citep{song2023moviechat, streamchat, longdocbench}, these approaches fail to fully capture the complexities of long-term memory in decision-making. Many existing QA benchmarks either rely on multiple-choice formats, which enable guessing, or require extensive human annotation to curate question-answer pairs that require models to attend to long video sequences. In contrast, we propose embodied memory tasks—such as \textit{``navigating to a soft toy that you did not rearrange yesterday"} or \textit{``find the second object picked up earlier"} that demand fine-grained recall and multi-hop reasoning over past observations. These tasks provide a more rigorous test of memory capabilities, requiring agents to track and respond to environmental changes rather than passively perceive static information.

In this paper, we introduce \ourbench, a benchmark designed to evaluate long-horizon memory in highly photorealistic simulated environments. Our benchmark consists of \textbf{\numtasks diverse navigation tasks}, which we categorize along various axes of memory requirements, that require both long-range temporal and spatial reasoning while discouraging heuristic shortcuts or random guessing. To ensure that performance is directly tied to memory utilization rather than simple perception, we curate scenarios where agents must recall past interactions to succeed. Additionally, our benchmark features dynamic environments - where agents modify the scene - making it essential to reason over changing contexts. Furthermore, \ourbench is procedurally extensible, allowing tasks to scale in complexity as VLMs improve, ensuring continued relevance as embodied AI progresses.

Unlike standard QA datasets that rely on subjective human annotations, our benchmark leverages photorealistic simulation to enable automatic evaluation of memory-based navigation. These tasks go beyond static recall since they require agents to perform complex spatio-temporal reasoning over past interactions. For instance, tasks such as \textit{``navigate to any object that you interacted with yesterday"} demand not only accurate memory retrieval, but also strategic decision-making to identify and navigate to the \textit{nearest} valid target from where the agent is currently positioned. 
In \ourbench, we introduce metrics that quantify task completion efficiency and not just successful recall.

Our key contributions are:
\vspace{-5pt}
\begin{enumerate}[leftmargin=2em, labelindent=0.5em, labelsep=0.5em]
    \item A benchmark for evaluating long-horizon memory in embodied decision-making, featuring \numtasks diverse navigation tasks in highly realistic indoor environments.
    \looseness-1 \item A comprehensive evaluation of VLM based high-level policies combined with low-level navigation policies, analyzing their performance and limitations in memory-intensive tasks.
    \item A systematic and extensible evaluation framework with metrics that enable future advancements in memory-efficient embodied agents. 
\end{enumerate}

\section{Related Work}
\label{sec:related}

In this section, we review related work on long-video question answering and embodied AI benchmarks that incorporate historical experience in their problem formulation. We defer the discussion of different approaches for augmenting policies with memory-like mechanisms to \cref{related_baselines}.

\subsection{Video QA benchmarks}
Recent analyses of video QA benchmarks~\citep{egoschema,atp} show that many tasks can be solved by attending to key frames, not full video sequences, suggesting that current benchmarks may favor keyframe selection over evaluating long-term memory. 

\vspace{-3pt}
\begin{table*}[t]
\centering
\caption{Comparison of \textit{\ourbench} with popular memory-focused embodied AI benchmarks.}
\vspace{-12pt}
\setlength{\tabcolsep}{3pt}
\resizebox{\textwidth}{!}{
\begin{tabular}{lccccccc}
\toprule
\textbf{Benchmark}                          & \textbf{Photorealism} & \textbf{Semantic Reasoning} & \textbf{Task Specification} & \textbf{Template Categories} & \textbf{\# Resulting Tasks} & \textbf{Memory Length ($\pi^*$)} & \textbf{Isolates Memory}  \\
\midrule
MemoryMaze~\citep{pasukonis2023evaluating}   & \xmark                & \xmark              & One-hot                & \xmark                & $1$                      & 500-1000 steps              & \xmark           \\
MemoryGym~\citep{pleines2023memory}          & \xmark                & \xmark              & One-hot                & \xmark                & $3$                      & 128-512 steps (Extensible)  & \xmark                   \\
MultiON~\citep{wani2020multion}              & \cmark                & \xmark              & One-hot                & \xmark                & $1$                        &  500 steps                 & \xmark               \\
SIF~\citep{2024sif}                          & \cmark                & \cmark              & Language                & $3$                   & $240$ (\texttt{val/test})                      &  \xmark \space (Uses semantic map)   & \xmark        \\
OpenEQA (Active)~\citep{OpenEQA2023}         & \cmark                & \cmark              & Language                & \xmark               & $1600$                     &  75 steps                   & \xmark              \\
GoatBench~\citep{khanna2024goatbench}        & \cmark                & \cmark              & Language + Image        & \xmark               & $1800-3600$ (\texttt{val})  & 500 steps (Static Env)                         & \xmark      \\ 
Excalibur~\citep{Zhu_2023_CVPR}              & \cmark                & \cmark              & Language                & $11$                & 21                          & Requires exploration         & \xmark           \\
    \hline
Ours                                        & \cmark                & \cmark              & Language                & $11$                & $\approx6000$ (\texttt{val})   & 500-3500 steps (Extensible)      & \cmark     \\
\bottomrule
\end{tabular}
}

\label{tab:benchmark_comparison}
\vspace{-12pt}
\end{table*}

Long-context evaluations for foundation models also focus on “needle-in-a-haystack” tasks, evaluating retrieval of specific details rather than integrating information across many segments~\citep{zhang2024longva}. While recent video benchmarks feature more challenging questions~\citep{egoschema, mvbench, wang2024novelqa, fu2024videomme, song2023moviechat, song2024milebench, li2023seed, rawal2024cinepile, yang2024thinking, mvlu, OpenEQA2023}, they adopt multiple-choice QA formats which are not suitable for evaluating fine-grained decision-making over long histories. They may also allow models to succeed through guessing and are not extensible since they require substantial human annotation effort.

\subsection{Embodied AI benchmarks}

Various benchmarks have been proposed to evaluate multi-modal language models in embodied settings where agents must leverage their experience in environments to improve future decision-making~\citep{wani2020multion}. 
Many focus on embodied question answering (EQA), such as HM-EQA~\citep{exploreeqa2024}, which requires active exploration, and Open-EQA~\citep{OpenEQA2023}, which tests VLMs in a passive setting with pre-collected episode histories.

Notably, Open-EQA found that a \emph{blind} LLM baseline—one that ignores visual input—still performed well, suggesting that multiple-choice evaluation can introduce biases and reduce the need for genuine memory recall.
Other benchmarks study the situated instruction following setting where agents must gather information to interpret ambiguous instructions \citep{seqa,ma2022sqa3d,2024sif}. However, many allow agents to sidestep memory challenges by precomputing relevant objects and locations based on recent history, effectively turning potentially memory-intensive tasks into structured instruction-following problems. Memory-Maze and Memory-Gym~\citep{pasukonis2023evaluating,pleines2023memory} focus on partially observable decision-making with history-based state modeling, but their simplistic visuals and small state spaces limit their relevance to complex real-world memory settings.

In \cref{tab:benchmark_comparison}, we compare our proposed benchmark to the most related benchmarks within embodied AI and RL research with respect to the following desired properties:
\vspace{-5pt}
\begin{enumerate}[leftmargin=2em, labelindent=0.5em, labelsep=0.5em]
    \item \textbf{Photorealism}: Measures the visual complexity of environments, affecting the meaningful evaluation of foundation models trained on real-world data.
    \item \textbf{Scope of reasoning needed over memory}: Distinguishes between benchmarks that require recalling diverse semantic information v.s. a narrow set of attributes (e.g. object locations).
    \item \textbf{Task Specification}: Evaluates whether tasks are defined using flexible language-based descriptions, enabling richer memory challenges.
    \item \textbf{Memory horizon length}: Adapting the definition from~\citet{ni2023when}, this measures the temporal scope over which memory must be maintained for optimal task performance. Unlike static benchmarks requiring localized frame recall, dynamic environments demand reasoning over events distributed across extended observation sequences. This metric becomes difficult to quantify when tasks require substantial exploration phases.
    \item \textbf{Memory isolation}: Whether benchmark performance isolates memory evaluation from confounding factors, particularly exploration requirements, which represent orthogonal challenges to memory recall and reasoning.

\end{enumerate}

Our benchmark isolates memory evaluation by avoiding confounding related to exploration and focuses on careful and extensible task curation with automated metrics, requiring agents to recall and reason over long-horizon environmental observations.
\section{\ourbench}
\label{sec:method}

In this section, we introduce our embodied memory benchmark, \ourbench, built on top of the \textbf{Habitat simulator}~\cite{habitat19iccv}. Habitat offers photorealistic visuals, diverse household scenes, and object-rich environments, making it a practical and \textbf{transferable testbed} for evaluating memory-augmented embodied agents. Since many vision-language models (VLMs) are pretrained on real-world image distributions, they can directly interpret observations from Habitat, enabling realistic and scalable evaluation of long-term decision-making.

We first describe the overall structure of \ourbench and the suite of tasks it comprises, then provide implementation details.

\subsection{Benchmark Design and Tasks}
\label{sec:instructions}
The goal of \ourbench is to evaluate an agent’s ability to retain and utilize past experiences to efficiently complete future tasks. To achieve this, we design a two-phase setup: an \textit{experience collection phase}, where scripted oracle agents autonomously gather experiences in the environment, and an \textit{interaction phase}, where agents are evaluated on their ability to recall and act upon that history. This decomposition isolates memory from exploration, enabling standardized evaluation.

During \textbf{experience collection}, oracle agents pick up and place objects onto designated receptacles (e.g., a table, chair, or bed), introducing meaningful state changes into the environment. Episodes span 400–3500 frames with 2–11 pick-and-place interactions (see \cref{appendix:experience_collection}). Once collection ends, the \textbf{interaction phase} begins: the agent is given access to its recorded history (images, pose, past actions) and tasked with following instructions that require integrating long-term memory into decision-making.

Tasks are instantiated from a set of \textbf{templated instructions} designed to probe diverse memory challenges. These cover \textbf{spatial}, \textbf{temporal}, and \textbf{semantic} reasoning: e.g., recalling which object was moved, when it was moved, or how it was interacted with; identifying rooms visited or farthest locations; and reasoning over attributes such as shape, color, or material. We also introduce \textbf{time-usage memory} tasks (e.g., longest room exploration, object with longest manipulation). A smaller subset of \textbf{attribute-based tasks} involves referring to objects indirectly via observed properties (e.g., ``go to the striped object you interacted with previously''), helping distinguish methods that can flexibly reconstruct fine-grained visual details from those relying on static summaries. In total, 60 templates are used to generate scene-specific task instances (see \cref{tab:all_tasks}).

\textbf{Single vs. Multiple Goals.}
Most tasks involve single-goal instructions, but some require reaching multiple destinations in sequence. Even single-goal tasks can be challenging, demanding agents recall what was \emph{not} interacted with or trace object movements across rearrangement states.

\looseness=-1
\textbf{Extensibility.}
The task suite is procedurally extensible: adjusting the complexity of experience collection (e.g., number of interactions) scales the temporal reasoning space and memory dependencies. We further group tasks into interpretable categories for analysis (see \cref{tab:task_categories}), making \ourbench one of the most visually complex and extensible embodied memory benchmarks to date.

\begin{table*}[htb]
    \centering
    \small  %
    \caption{Task categories with example prompts and challenges. See \cref{appendix:qual_examples} for more examples.}
    \vspace{-10pt}
    \renewcommand{\arraystretch}{1.3}
    \begin{tblr}{
        width = \linewidth,
        colspec = {X[0.04,l,m]X[0.9,l,m]X[3.0,l,m]X[2.1,l,m]},
        hline{1,2,13},
        row{1} = {font=\bfseries, bg=lightgray!30},
        row{even} = {bg=lightgray!10},
        cell{1}{2,3,4} = {font=\bfseries, c},
        column{1} = {colsep=7pt},
        column{2} = {colsep=1pt},
        column{3} = {colsep=1pt},
        column{4} = {colsep=1pt},
    }
        & Task Category & Example Instructions & Memory Requirement \\
        \SetCell[r=6]{m,bg=lightgray!10, halign=c}{\rotatebox{90}{Spatial}}
        & Object Recall & Navigate to a \textit{\{category\}}.  & 
        Object was previously seen or needs exploration. \\
        & Interaction & Navigate to any object that you did not interact with yesterday. & 
        Differentiate between interacted \& uninteracted objects. \\
        & Conditional Interaction & Navigate to a \textit{\{receptacle\_category\}} you picked an object from. & 
        Object categories, object-receptacle relationships. \\
        
        & Object Attribute & Navigate back to a \textit{\{target\_color\}} colored object that you interacted with yesterday. & 
        Object attributes other than category. \\
        
        & Spatial Relationship & Navigate to the object which you interacted with which is the farthest from your current location. & 
        Spatial relationships to objects. \\
        
        & Room Visitation &
        Navigate to a room that you did not visit yesterday. & 
        Room boundaries and past interactions. \\
        \SetCell[r=3]{m,bg=white,halign=c}{\rotatebox{90}{Temporal}}
        & Interaction Order & Navigate to the object you interacted with immediately after \textit{\{object\_category\}}. & 
        Sequence of objects interactions. \\
        & Time-Based & Navigate to the object that you interacted with at \textit{\{HH:MM\}} yesterday. & 
        Time of interaction \& final position of objects. \\
        & Duration Tracking & Navigate to the object which took the longest time to rearrange. & 
        Duration of interactions. \\
        \SetCell[r=2]{m,bg=lightgray!10,halign=c}{\rotatebox{90}{Multi-Goal}}
        & Unordered Revisitation & Revisit all the receptacles you picked objects from yesterday. & 
        Past locations of multiple objects and best visitation order. \\
        & Ordered Revisitation & Revisit all the objects you interacted with yesterday in specific order. & 
        Similar to unordered revisitation but with strict sequencing. \\
    \end{tblr}
    \label{tab:task_categories}
    \vspace{-12pt}
\end{table*}

\subsection{Implementation details}
\label{sec:implementation_details}

\textbf{Simulation Setup and Dataset Composition.}
We build \ourbench using the Habitat simulator~\citep{habitat19iccv,szot2021habitat} and scenes from the Habitat Synthetic Scenes Dataset (HSSD)~\citep{hssd}. Our benchmark uses 107 scenes from the training split and 30 from the validation split, with 1,478 training and 100 validation episodes. We populate these scenes with the same object sets used in OVMM~\citep{yenamandra2023homerobot}.
Across these episodes, we include 839/247 object instances spanning 84/72 distinct categories in the \texttt{train/val} splits, respectively. Objects are rearranged during the experience collection phase to induce temporally grounded memory challenges. Additionally, agents interact with 17/16 distinct receptacle categories in the \texttt{train/val} splits (e.g., tables, chairs, beds).
To support attribute-based tasks (see \cref{tab:task_categories}), we generate semantic descriptions of object instances using GPT-4o. We prompt the model with multi-view images of each object to extract attributes such as shape, color, material, and functionality. These descriptions are used to populate the object attribute task templates (5 total), which are limited to the validation set and undergo manual quality review. In total, our benchmark includes 79,213 training and 5,876 validation task instances. More details in~\cref{sec:dataset_creation}.

\textbf{Agent Setup.} We adopt the Hello Robot Stretch embodiment~\citep{kemp2022design} in our simulation. 
The agent is equipped with RGB and depth sensors (resolution $640 \times 480$) and a GPS+Compass sensor. The agent can execute one of four discrete navigation actions: \texttt{MOVE\_FORWARD} ($0.25$m), \texttt{TURN\_LEFT} / \texttt{TURN\_RIGHT} (by $10^\circ$) and a \texttt{STOP} action. During the experience collection phase, the agent can also manipulate its arm joints (extension, lift, yaw, pitch, roll) and activate a \texttt{manipulation\_mode} for pick-and-place actions, which rotates the base ($90^\circ$) and directs the head camera toward the gripper.

\textbf{Procedural Generation and Verification using PDDL.} We operate over a large set of diverse task instructions (see ~\cref{tab:task_categories}) by encoding them into templates using the PDDL specification system ~\citep{aeronautiques1998pddl,szot2024large}.
Specifically, each template specifies \textit{entities} (objects and receptacles) with \textit{properties} such as interaction order, target color, and start/goal receptacle. 
These entities are combined using \textit{predicates} that map to pythonic functions which can be evaluated against any simulator state. 
\textit{Goal conditions} compose these predicates with \textit{quantifiers} (e.g., \texttt{EXISTS}, \texttt{FOR\_ALL\_UNORDERED}) to express complex logical relationships between multiple predicates.
The defined instructions templates are \textit{procedurally} expanded by sampling valid entity bindings based on the entity properties which serve as constraints.
The bindings are finally substituted into the goal conditions to generate the \ourbench dataset and enable scalable, automated verification for each episode without hand-coding low-level rules.
\section{Experiments}

\label{experiments}
We evaluate several baseline approaches on \ourbench and assess their ability to complete memory-intensive tasks using a combination of vision-language reasoning and navigation. Below, we describe the baseline architecture, implementation details, and evaluation metrics.

\subsection{Evaluated Approaches}
\label{sec:hierarchical_baseline}

To establish strong baselines, we adopt a hierarchical architecture (\cref{fig:ourbench_agent}) comprising two modules: (1) a high-level goal selection module, which processes the interaction history to select a sequence of goal frames, and
(2) a low-level navigation policy, which executes actions to reach the selected frame. This architecture enables reasoning over long-horizon multimodal experience while delegating fine-grained control to a learned policy. Inspired by prior work~\citep{xu2024mobility}, we avoid directly predicting actions using VLMs due to their poor performance in continuous control, and instead restrict them to high-level goal prediction.

\subsubsection{High-Level Goal Selection}
\looseness=-1 
\textbf{Video LM Agent.}
This agent receives the full interaction video along with a task instruction. Each frame in the video is annotated with its index and the time of day at which it was captured. The agent is tasked with predicting the index of a frame from the video that corresponds to a viable goal state. Specifically, the predicted frame should represent a viewpoint from which, if the agent were to navigate back to the same camera position, it could successfully complete the task. Although the model outputs a single frame index, selecting the correct one often requires reasoning over multiple temporally scattered observations, including frames where the agent may not be actively interacting. Since different VLMs are capable of handling different numbers of frames in context, we subsample the frames passed to each model based on what empirically performs best. We evaluate both proprietary (Gemini-2.0-Flash~\citep{team2024gemini}, GPT-4o) and open-source (Qwen2.5-VL~\citep{qwen25vl}, Gemma-3~\citep{team2025gemma}, GLM-4.1V-Thinking~\citep{hong2025glm}) vision-language models capable of processing long video contexts. See \cref{appendix:gemini_baseline,appendix:qwen_gemma_baseline,appendix:gpt_baseline} for more details.

\looseness=-1 

\textbf{Textual Memory Agent.} Similar to~\citep{2024sif,remember}, we also consider an explicit, memory-building baseline which breaks down the video frames into chunks and uses a VLM to generate relevant textual summaries for each video chunk. After this guided ``captioning" of the entire video sequence, an LLM selects goal frame indices by reasoning over the generated textual descriptions. See \cref{appendix:text_agent} for more details.

\textbf{Supervised Fine-Tuning (SFT).} We additionally explore a supervised fine-tuning approach where the high-level VLM is trained to predict goal frame indices, given the task prompt and interaction history. Since multiple frames may validly represent the goal (e.g., several consecutive frames showing an object being placed), we supervise the model on the full list of acceptable target frames. At test time, a single frame is sampled from the list of predictions and is considered correct if it matches any of these valid frames. This setup enables the model to learn task-specific cues from examples and serves as a stronger baseline than zero-shot prompting, particularly for tasks requiring subtle temporal grounding.
We provide additional details in \cref{appendix:sft_baseline}.

\subsubsection{Low-Level Navigation}

To execute the high-level plan, we use a low-level policy that takes the selected goal frame and outputs discrete actions: \texttt{MOVE-FORWARD}, \texttt{TURN-RIGHT}, \texttt{TURN-LEFT}, and \texttt{STOP}. Our primary policy is the LSTM-based controller from OVRLv2~\citep{yadav2023ovrlv2}, trained to reach image goals from egocentric RGB-D observations (see~\cref{sec:imagenav_details}) (ImageNav). We also evaluate a mapping-based policy that deterministically navigates to the location corresponding to the goal frame using a global map, without relying on learned visual features (see~\cref{sec:mapper_details}).

\textbf{Solvability.} The action space of our high-level baseline is constrained to frames observed during experience collection, which is insufficient for some tasks. 
For example, in the task \textit{“Navigate to a room that you did not visit yesterday”} all frames correspond to previously visited locations, making it impossible to complete the instruction.
To establish an upper bound on task difficulty, we quantify the percentage of episodes that could theoretically be solved through optimal frame selection from interaction history, assuming perfect navigation (teleportation) to the selected waypoint. 
This upper bound is the dashed line (\texttt{Oracle}) in \cref{fig:ourbench_main_results}.

\subsection{Evaluation Metrics}
\label{sec:eval_metrics}

We benchmark agent performance on the \ourbench tasks using \textbf{Low Level Success Rate} (\texttt{LL-SR}), assessing whether the agent finds the right target entities, and \textbf{Low Level Success-weighted-by-Path-Length} (\texttt{LL-SPL}), evaluating if the agent selects and navigates to the optimal entity using the shortest possible path. Since we use a hierarchical baseline, we additionally introduce a \textbf{High-Level Policy Success Rate}  (\texttt{HL-SR}), and \textbf{High-Level Policy SPL} (\texttt{HL-SPL}) which measures the accuracy and efficiency of the high-level policy in selecting the correct and closest goal frames for navigation. To further analyze failure modes, we also report two relaxed variants: \textbf{Distance-to-Goal-Only SR} (\texttt{DTG-SR}), which relaxes the requirement that the target object must be visible in the selected frame, and \textbf{Semantic Coverage SR} (\texttt{SC-SR}), which ignores spatial proximity.
Additional metrics, success criteria, and threshold specifics are provided in~\cref{app:evaluations}.

\section{Results}

\begin{figure*}[t]
  \centering
  \includegraphics[width=\linewidth]{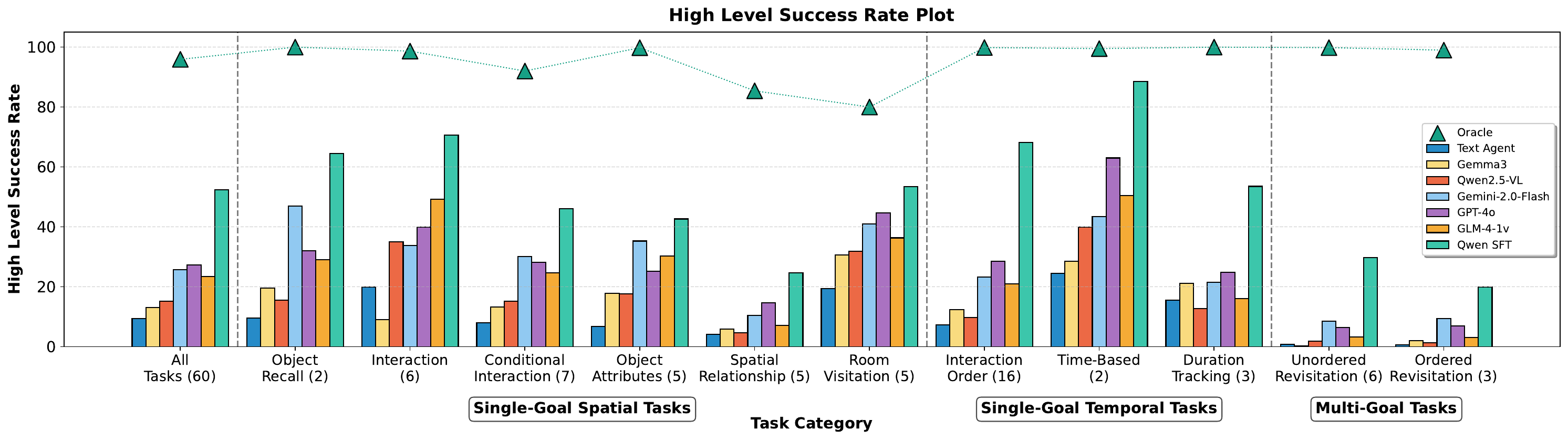} 
  \vspace{-20pt}
  \caption{\ourbench Task Performance: Proprietary VLMs such as \texttt{Gemini-2.0-flash} and \texttt{GPT-4o} show little success across most \ourbench task categories. Agents struggle especially on multi-goal tasks where multiple subgoals must be achieved sequentially from the interaction videos collected during Phase 1. The \texttt{Qwen-SFT} baseline (see \cref{appendix:sft_baseline}), trained to predict ground-truth frame indices, performs best but still reaches only $\approx 50\%$ on high-level tasks.}
  \vspace{-12pt}
\label{fig:ourbench_main_results}
\end{figure*}

In this section, we analyze the performance of the previously outlined baselines on \ourbench tasks, presenting results grouped according to the task categories described in~\cref{sec:instructions}.

\textbf{State-of-the-Art VLMs struggle on the \ourbench benchmark.} We first evaluate different VLMs on the high-level task and report success rates in~\cref{fig:ourbench_main_results}, observing consistently low performance across all task subsets (see \cref{tab:goal_success} for numbers). GPT-4o achieves the highest success at 27.3\%, followed by Gemini-2.0-Flash at 25.7\%, and GLM-4.1V-Thinking at 23.5\%. The non-reasoning open-source models Qwen2.5-VL and Gemma-3 score between 13\% and 16\%. The text-based agent performs the worst overall, underscoring that textual memory representations alone are insufficient for disambiguating visual goals or reasoning over complex visual histories. Notably, the supervised fine-tuned (SFT) variant outperforms all frozen VLMs, achieving an average improvement of 25\% across tasks, with considerable improvement on temporal and multi-goal tasks.

In single-goal tasks, all methods struggle particularly with the spatial relationship subset, with GPT-4o achieving only 14.5\% success. These tasks require reasoning about relative distances between entities and the agent, an ongoing challenge noted in prior work~\citep{yang2024thinking}. 
In contrast, room identification tasks yield the highest performance, likely due to the small number of distinct rooms and the lenient success criteria of reaching any location within the correct room. 
Performance on conditional interaction and object attribute tasks is comparable, suggesting that current VLMs handle both category-based queries (e.g., ``apple") and attribute-based descriptions (e.g., ``red food") with similar effectiveness.

Temporal tasks pose additional challenges to the existing VLMs, with models consistently failing to identify the correct order of interactions, since it requires multi-hop reasoning. 
Duration tracking is similarly difficult as it requires inferring elapsed duration based on specific points in the video.
In contrast, for the time-based tasks, we observe that baselines perform better, as these tasks only require single frame retrieval based on the timestamps in the question, without additional reasoning.

\textbf{Poor performance on multi-goal tasks.} From~\cref{fig:ourbench_main_results}, we observe that all VLM baselines achieve near-zero performance on most multi-goal tasks. This indicates that VLMs struggle to track and recall multiple object–receptacle interactions. In our multi-goal evaluation, we employ a single-shot prompt, requiring the high-level VLM agent to predict all target frame indices in one response, primarily to avoid repeated inferences over long sequences given the large number of frames. Qualitatively, we find that VLMs often fail to predict even the correct number of targets needed to solve a task. While our trained Qwen model performs somewhat better, reaching around 20\% accuracy, it still falls far short of the attainable 99\% success rate on this task.

\begin{figure*}[t]
    \centering
    \small
    \begin{subfigure}{0.245\textwidth}
        \centering
        \includegraphics[width=\linewidth]{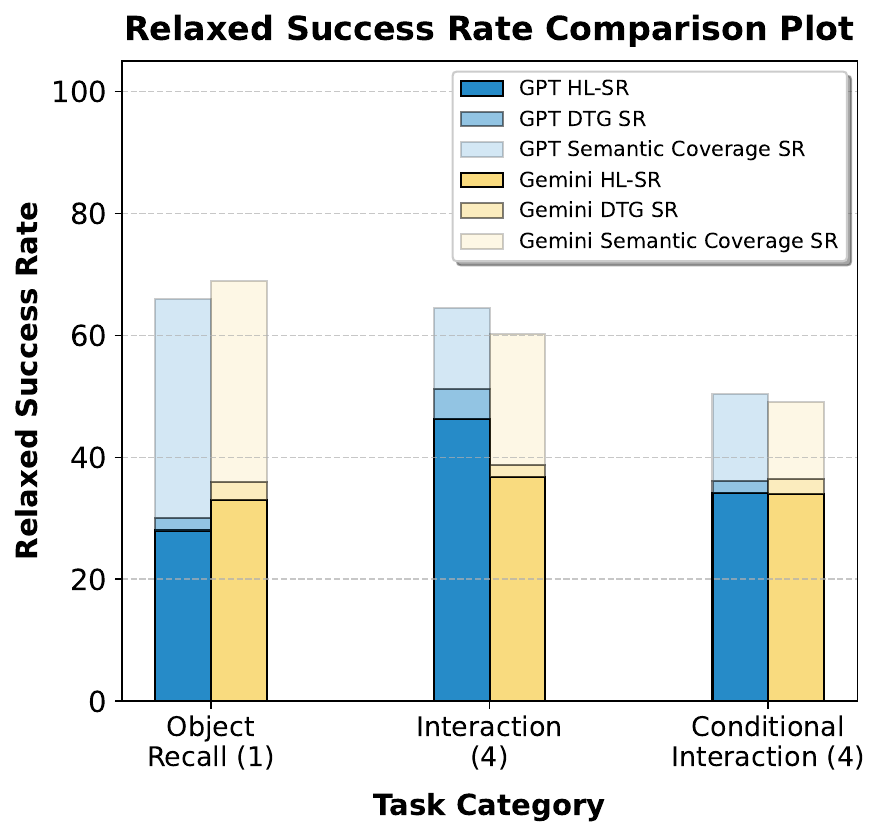}
        \vspace{-18pt}
        \caption{\scriptsize Relaxed metrics: Recep.}
        \label{fig:relaxed_metrics_recep}
    \end{subfigure}
    \begin{subfigure}{0.245\textwidth}
        \centering
        \includegraphics[width=\linewidth]{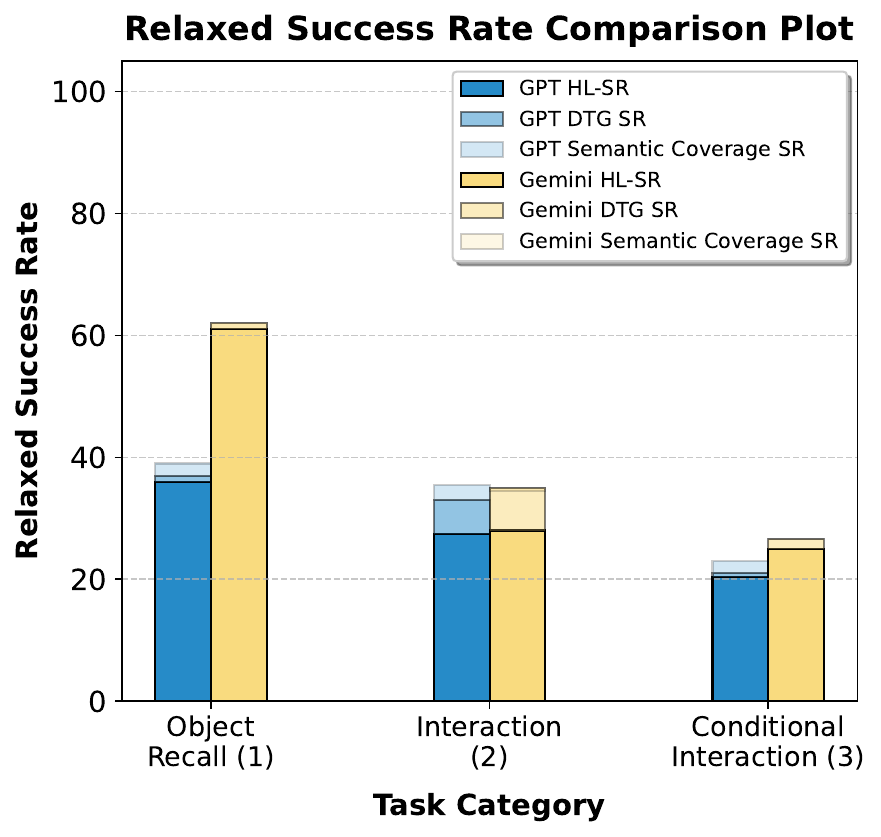}
        \vspace{-18pt}
        \caption{\scriptsize Relaxed metrics: Objects.}
        \label{fig:relaxed_metrics_obj}
    \end{subfigure}
    \begin{subfigure}{0.245\textwidth}
        \centering
        \includegraphics[width=\linewidth]{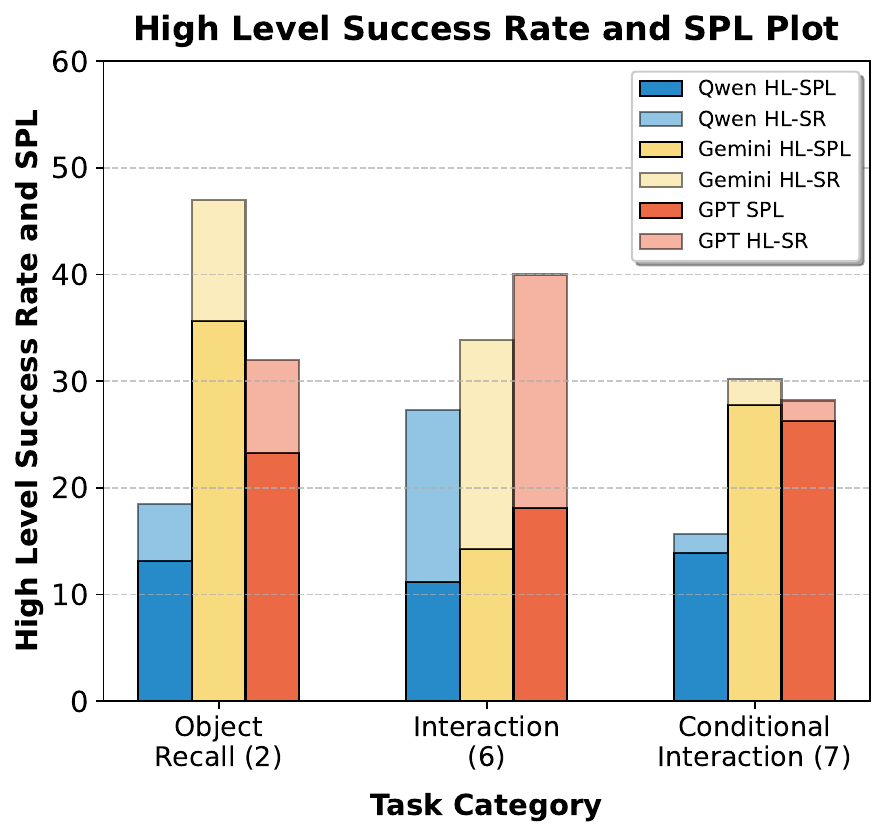}
        \vspace{-18pt}
        \caption{\scriptsize HL-SR vs HL-SPL.}
        \label{fig:high_level_spl}
    \end{subfigure}
    \begin{subfigure}{0.245\textwidth}
        \centering
        \includegraphics[width=\linewidth]{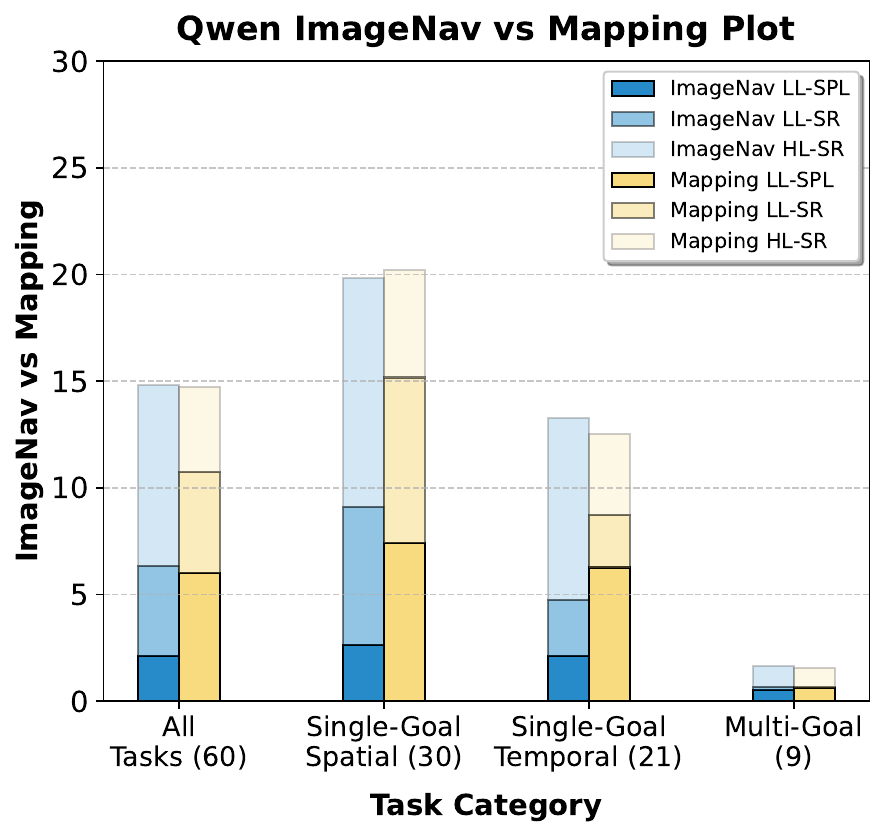} %
        \vspace{-18pt}
        \caption{\scriptsize Hierarchical policy.}
        \label{fig:low_level_sr}
    \end{subfigure}
    \vspace{-18pt}
    \caption{\textbf{(a, b) Relaxed metrics for object and receptacle tasks.} Models show substantial gains on receptacle-based tasks under relaxed metrics, whereas tasks targeting small objects exhibit little improvement. \textbf{(c) Success Rate vs. SPL.} For tasks where the VLM may select among multiple correct frames, all models sometimes choose suboptimal ones, reducing the \texttt{HL-SPL} metric. \textbf{(d) Hierarchical policy on \ourbench.} When Qwen is paired with an ImageNav or Mapping agent, overall SR and SPL drop markedly (as reflected in \texttt{LL-SR/LL-SPL}).}
    \vspace{-12pt}
\end{figure*}

\textbf{VLMs can identify target receptacles but not precisely localize them.}
We analyze VLM performance on tasks where the valid entities are receptacles versus where they are objects. As shown in~\cref{fig:relaxed_metrics_recep}, the \texttt{Semantic Coverage-SR (SC-SR)} metric (see~\cref{app:evaluations}) is substantially higher for receptacle-based tasks. This indicates that VLMs often select frames where the receptacle is visible, but the agent remains far from it. In contrast, distance-threshold success (\texttt{DTG-SR}) improves negligibly, suggesting that very few cases exist where the agent is close to a receptacle but not facing it. For smaller objects, no such improvement is observed (see~\cref{fig:relaxed_metrics_obj}), since small objects are rarely visible from longer distances. Taken together, these results highlight a key limitation of current VLMs: they can semantically detect large targets but fail to construct spatially grounded representations necessary for precise localization.

\textbf{VLMs struggle with identifying the closest entity that solves the task.} In Fig. \ref{fig:high_level_spl} we report the \texttt{HL-SPL} metric (\cref{sec:eval_metrics}) which captures the efficiency with which the high level VLM-based modules select goal frames.
During the experience collection phase, the agent can interact with multiple object instances of the same category due to which a task such as \textit{``Navigate to a squeezer you interacted with"} could have multiple valid entities that the agent can navigate to.
We observe that for such tasks there is a wide gap between \texttt{HL-SR} and \texttt{HL-SPL} (upto $50\%$ in case of interaction-based tasks) for all model variants.
This suggests that although the VLMs can correctly recognize a valid entity to solve the task, they still fail to perform a fine-grained spatial analysis of the multiple valid entities being rearranged dynamically in the video. Thus, they cannot spatially locate the valid entity nearest to its current position. 

\looseness=-1 
\textbf{Policy decomposition causes performance degradation.} To evaluate the hierarchical agent introduced in~\cref{sec:hierarchical_baseline}, we use \texttt{Qwen2.5-VL-7B} as the high-level VLM module and pair it with either an ImageNav policy or a mapping-based policy as the low-level agent. The low-level policy is triggered only when the high-level VLM correctly identifies a subgoal. As shown in~\cref{fig:low_level_sr}, the \texttt{Low-Level-SR (LL-SR)} metric drops sharply across all \ourbench tasks for the ImageNav policy. We attribute this decline to a distribution shift in the visual goals provided at evaluation. During training, the policy sees randomly sampled start–end goal poses, whereas VLM-selected goals often correspond to pick–place routines. These frames largely contain background walls or partial receptacle surfaces, offering limited visual cues for navigation. By contrast, the mapping-based navigation policy is more robust, suffering only a 25\% relative drop in SR. However, the SPL of both methods remains in the single digits, underscoring the importance of \ourbench in exposing these shortcomings and motivating the development of more reliable memory and navigation strategies. 

\begin{figure*}[ht] %
    \centering
    \vspace{-8pt}
    \begin{subfigure}[b]{0.49\linewidth}
        \centering
        \includegraphics[width=\linewidth]{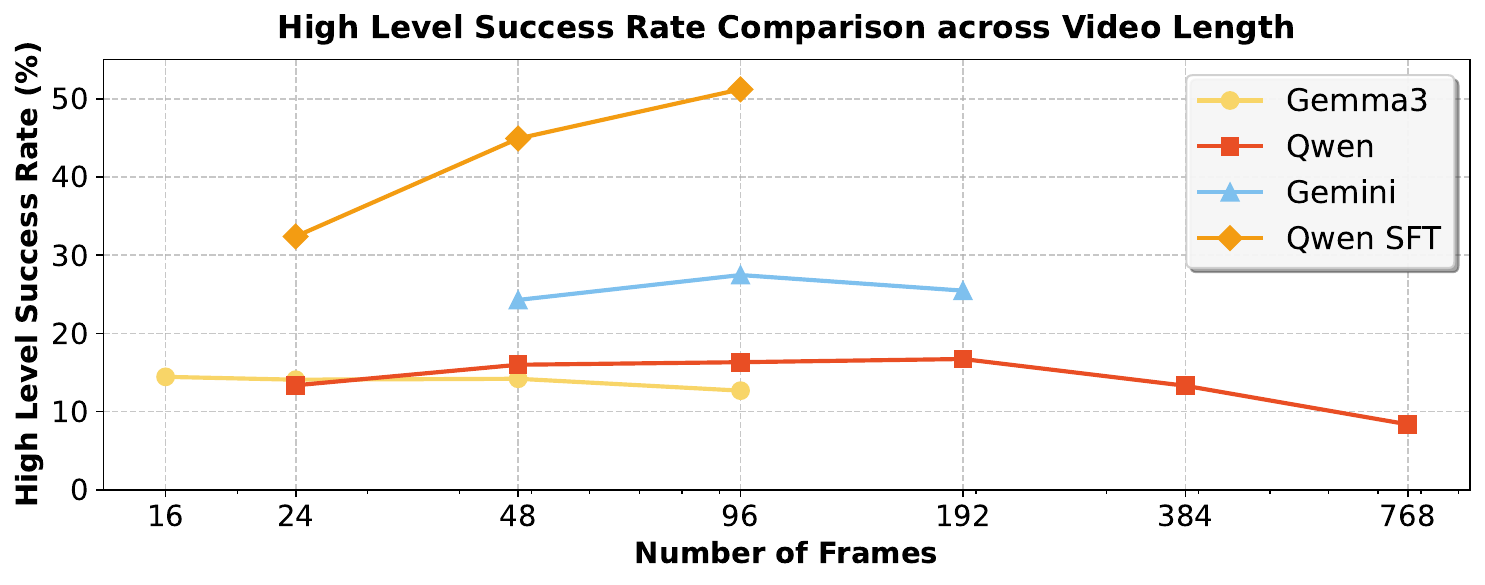}
        \vspace{-10pt}
        \caption{Comparison across different video lengths.}
        \label{fig:model_checkpoint_comparison}
    \end{subfigure}
    \hfill
    \begin{subfigure}[b]{0.49\linewidth}
        \centering
        \includegraphics[width=\linewidth]{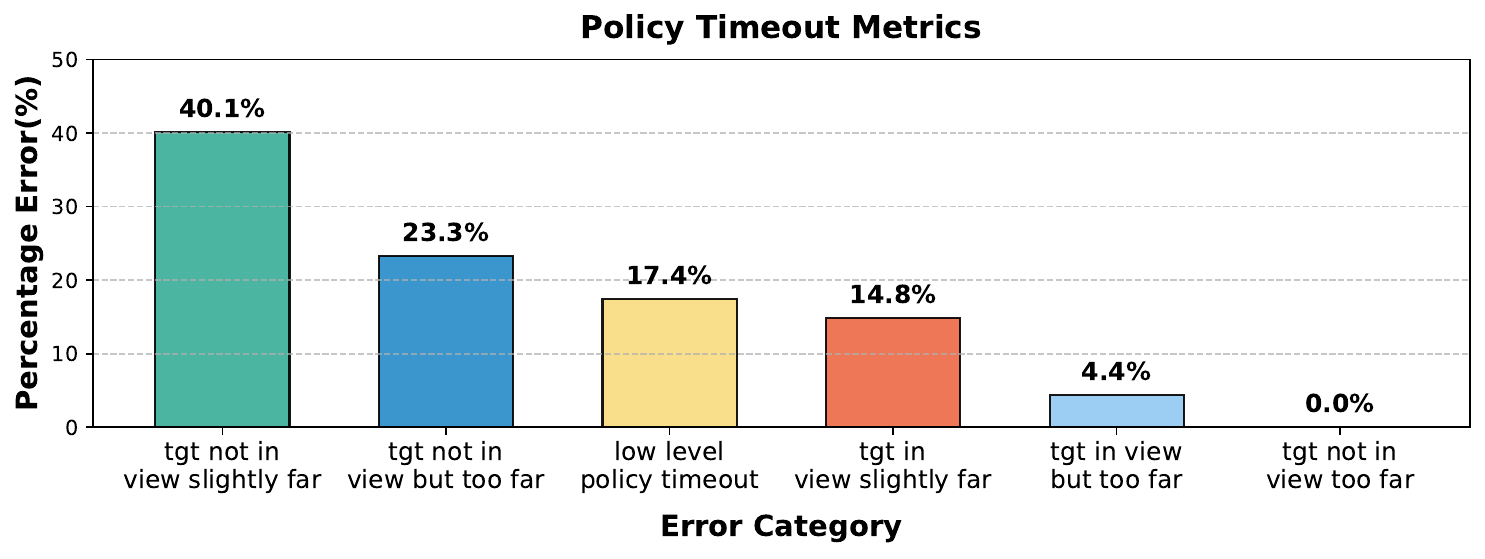}
        \vspace{-10pt}
        \caption{ImageNav Policy Errors}
        \label{fig:failure_imagenav}        
    \end{subfigure}
    \vspace{-8pt}
    \caption{\textbf{(a)} Model performance at different number of subsampled video frames. \textbf{(b)} Reasons for ImageNav Policy Errors.}
    \vspace{-8pt}
    \label{fig:combined_imagenav}
\end{figure*}

\textbf{Low level navigation policy failures.} From ~\Cref{fig:failure_imagenav}, we observe that the low-level image-goal navigation policy can fail due to invoking the termination (stop) action incorrectly or a navigation routine timeout (max budget of $1000$ steps including the data collection phase). To further understand this behavior, we analyze if the navigation policy actually looks at the target entity when the stop action is invoked. Interestingly, we find that in several instances the target entity is visible from a slightly larger distance threshold (2.0m) from the chosen task success threshold (1.0 m, in case of receptacles). This often happens when VLMs select goal frames in which the target entity is visible in the 2D image but not necessarily closest in 3D space, and thus the goal frame coordinates are slightly further away than the threshold distance.

\textbf{Leveraging additional frames.}
We study the effect of input length on high-level performance by evaluating agents on subsampled interaction videos of varying sizes (\cref{fig:model_checkpoint_comparison}). This allows us to test whether models benefit from longer histories or merely pick up sparse visual cues. Interestingly, frozen VLMs, although capable of processing long contexts, show little to no improvement with more frames. Their performance often degrades at higher frame counts, suggesting difficulty in extracting relevant signals from densely packed, unfiltered inputs.

In contrast, the fine-tuned model (Qwen SFT) achieves clear gains when trained with longer videos. This shows that with appropriate supervision, models can move beyond shallow matching to exploit richer temporal structure in interaction histories. Nonetheless, even with full finetuning on longer (\textit{subsampled}) videos, a large gap remains relative to maximum achievable performance (see~\cref{app:task_solvability}). This underscores the need for future methods that enable VLMs to extract and store in a compressed manner (i.e. via memory mechanisms) task-relevant information more effectively from long video sequences within the context window.

\begin{wraptable}{r}{0.5\textwidth}
    \centering
    \small
    \vspace{-1.5em} %
    \resizebox{1\linewidth}{!}{%
    \begin{tabular}{lcc}
        \toprule
        \textbf{Method} & \textbf{MCQ Accuracy} (\(\uparrow\)) & \textbf{Numerical Accuracy} (\(\uparrow\)) \\
        \midrule
        Zero-Shot            & 33.69 & 21.47 \\
        \hdashline
        Video-R1-Only SFT   & 33.33       & 32.67 \\
        \rowcolor{gray!15} Video-R1 + \ourbench SFT       & 35.06        & 32.35  \\
        \hdashline
        Video-R1-Only GRPO      & 33.57                 & 33.9 \\
        \rowcolor{gray!15} Video-R1 + \ourbench GRPO     & \textbf{35.42}                 & \textbf{34.82} \\
        \bottomrule
    \end{tabular}
    }
    \vspace{-0.5em}
    \caption{\texttt{VSI-Bench} results. Models trained with \ourbench show consistent gains.}
    \vspace{-1.0em} %
    \label{tab:vsi_bench_results}
\end{wraptable}

\looseness=-1
\textbf{Cross-Benchmark Results.} In this experiment, we explore whether the spatiotemporal memory capabilities enhanced by training on simulator-based \ourbench data can translate to real world settings. 
To evaluate this, we use VSI-Bench ~\citet{yang2024thinking}, a challenging benchmark specifically designed to measure spatiotemporal reasoning in VLMs on egocentric real-world videos.
We first augment the \texttt{Video-R1-CoT-165k} dataset introduced in~\citet{feng2025video} with $40000$ samples from the \ourbench dataset.
For effective co-training, we construct chain-of-thought traces for the entire training split of the \ourbench dataset (see \cref{app:cot_generation}).
We provide specifics about training and evaluation in ~\cref{appendix:cross_benchmark_exps}.
We then evaluate the trained model on VSI-Bench, showing the results in ~\cref{tab:vsi_bench_results}. We observe that co-training with \ourbench improves the SFT baseline performance by $1.7\%$ (MCQ accuracy). 
Furthermore, RL-finetuning with GRPO using combined data from the \ourbench dataset improves performance with a $1.85\%$ increase over the Video-R1-Only RL checkpoint (MCQ accuracy). 
Overall, the consistent gains obtained from co-training point to a sim-to-real transfer effect, suggesting that \ourbench can augment scarce real-world data for advancing spatiotemporal understanding in VLMs.

\textbf{Qualitative Examples.} We present multiple example responses from various VLMs in ~\cref{appendix:qual_examples}.
Overall we observe that closed models perform better than open-source counterparts on single-goal tasks (see \cref{fig:spatial_ex_1,fig:temporal_ex_1,fig:temporal_ex_2}) but all models suffer on multi-goal tasks (see ~\cref{fig:multigoal_ex}).
We also provide example agent video trajectories at 
\url{https://findingdory-benchmark.github.io/}.
\section{Conclusion}
We introduced \ourbench, a benchmark for evaluating long-horizon memory in embodied agents within a highly photorealistic indoor simulation. 
Our findings reveal limitations in VLM context scaling, particularly in handling long observation sequences required for memory-intensive tasks. 
\ourbench provides informative metrics that disentangle different aspects of memory reasoning, allows for procedural task generation, and efficiency-focused failure analysis beyond simple frame selection accuracy. 
Its extensibility allows for longer temporal dependencies, supporting research on context scaling and memory efficiency in multimodal models.
Our experiments also highlight challenges with hierarchical policies motivating the need for better architectures that closely integrate long-context VLM capabilities with generalist embodied policies.

\newpage

\bibliography{iclr2026_conference}

\begin{thebibliography}{53}
\providecommand{\natexlab}[1]{#1}
\providecommand{\url}[1]{\texttt{#1}}
\expandafter\ifx\csname urlstyle\endcsname\relax
  \providecommand{\doi}[1]{doi: #1}\else
  \providecommand{\doi}{doi: \begingroup \urlstyle{rm}\Url}\fi

\bibitem[gem()]{gemini-api}
Explore vision capabilities with the gemini api.
\newblock \emph{\url{https://ai.google.dev/gemini-api/docs/vision?lang=python}}.

\bibitem[Aeronautiques et~al.(1998)Aeronautiques, Howe, Knoblock, McDermott, Ram, Veloso, Weld, Sri, Barrett, Christianson, et~al.]{aeronautiques1998pddl}
Constructions Aeronautiques, Adele Howe, Craig Knoblock, ISI~Drew McDermott, Ashwin Ram, Manuela Veloso, Daniel Weld, David~Wilkins Sri, Anthony Barrett, Dave Christianson, et~al.
\newblock Pddl| the planning domain definition language.
\newblock \emph{Technical Report, Tech. Rep.}, 1998.

\bibitem[{Anwar} et~al.(2024){Anwar}, {Welsh}, {Biswas}, {Pouya}, and {Chang}]{2024arXivremembr}
Abrar {Anwar}, John {Welsh}, Joydeep {Biswas}, Soha {Pouya}, and Yan {Chang}.
\newblock {ReMEmbR: Building and Reasoning Over Long-Horizon Spatio-Temporal Memory for Robot Navigation}.
\newblock \emph{arXiv e-prints}, art. arXiv:2409.13682, September 2024.
\newblock \doi{10.48550/arXiv.2409.13682}.

\bibitem[Anwar et~al.(2024)Anwar, Welsh, Biswas, Pouya, and Chang]{remember}
Abrar Anwar, John Welsh, Joydeep Biswas, Soha Pouya, and Yan Chang.
\newblock Remembr: Building and reasoning over long-horizon spatio-temporal memory for robot navigation.
\newblock In \emph{8th Annual Conference on Robot Learning}, 2024.

\bibitem[Bai et~al.(2025)Bai, Chen, Liu, Wang, Ge, Song, Dang, Wang, Wang, Tang, Zhong, Zhu, Yang, Li, Wan, Wang, Ding, Fu, Xu, Ye, Zhang, Xie, Cheng, Zhang, Yang, Xu, and Lin]{qwen25vl}
Shuai Bai, Keqin Chen, Xuejing Liu, Jialin Wang, Wenbin Ge, Sibo Song, Kai Dang, Peng Wang, Shijie Wang, Jun Tang, Humen Zhong, Yuanzhi Zhu, Mingkun Yang, Zhaohai Li, Jianqiang Wan, Pengfei Wang, Wei Ding, Zheren Fu, Yiheng Xu, Jiabo Ye, Xi~Zhang, Tianbao Xie, Zesen Cheng, Hang Zhang, Zhibo Yang, Haiyang Xu, and Junyang Lin.
\newblock Qwen2.5-vl technical report, 2025.
\newblock URL \url{https://arxiv.org/abs/2502.13923}.

\bibitem[Batra et~al.(2020)Batra, Gokaslan, Kembhavi, Maksymets, Mottaghi, Savva, Toshev, and Wijmans]{batra2020objectnav}
Dhruv Batra, Aaron Gokaslan, Aniruddha Kembhavi, Oleksandr Maksymets, Roozbeh Mottaghi, Manolis Savva, Alexander Toshev, and Erik Wijmans.
\newblock Objectnav revisited: On evaluation of embodied agents navigating to objects.
\newblock \emph{arXiv preprint arXiv:2006.13171}, 2020.

\bibitem[{Buch} et~al.(2022){Buch}, {Eyzaguirre}, {Gaidon}, {Wu}, {Fei-Fei}, and {Niebles}]{atp}
Shyamal {Buch}, Crist{\'o}bal {Eyzaguirre}, Adrien {Gaidon}, Jiajun {Wu}, Li~{Fei-Fei}, and Juan~Carlos {Niebles}.
\newblock {Revisiting the ``Video'' in Video-Language Understanding}.
\newblock \emph{arXiv e-prints}, art. arXiv:2206.01720, June 2022.
\newblock \doi{10.48550/arXiv.2206.01720}.

\bibitem[{Chang} et~al.(2023){Chang}, {Gervet}, {Khanna}, {Yenamandra}, {Shah}, {Min}, {Shah}, {Paxton}, {Gupta}, {Batra}, {Mottaghi}, {Malik}, and {Singh Chaplot}]{goat}
Matthew {Chang}, Theophile {Gervet}, Mukul {Khanna}, Sriram {Yenamandra}, Dhruv {Shah}, So~Yeon {Min}, Kavit {Shah}, Chris {Paxton}, Saurabh {Gupta}, Dhruv {Batra}, Roozbeh {Mottaghi}, Jitendra {Malik}, and Devendra {Singh Chaplot}.
\newblock {GOAT: GO to Any Thing}.
\newblock \emph{arXiv e-prints}, art. arXiv:2311.06430, November 2023.
\newblock \doi{10.48550/arXiv.2311.06430}.

\bibitem[Feng et~al.(2025)Feng, Gong, Li, Guo, Wang, Peng, Wu, Zhang, Wang, and Yue]{feng2025video}
Kaituo Feng, Kaixiong Gong, Bohao Li, Zonghao Guo, Yibing Wang, Tianshuo Peng, Junfei Wu, Xiaoying Zhang, Benyou Wang, and Xiangyu Yue.
\newblock Video-r1: Reinforcing video reasoning in mllms.
\newblock \emph{arXiv preprint arXiv:2503.21776}, 2025.

\bibitem[Fu et~al.(2024)Fu, Dai, Luo, Li, Ren, Zhang, Wang, Zhou, Shen, Zhang, et~al.]{fu2024videomme}
Chaoyou Fu, Yuhan Dai, Yondong Luo, Lei Li, Shuhuai Ren, Renrui Zhang, Zihan Wang, Chenyu Zhou, Yunhang Shen, Mengdan Zhang, et~al.
\newblock Video-mme: The first-ever comprehensive evaluation benchmark of multi-modal llms in video analysis.
\newblock \emph{arXiv preprint arXiv:2405.21075}, 2024.

\bibitem[Guo et~al.(2025)Guo, Yang, Zhang, Song, Zhang, Xu, Zhu, Ma, Wang, Bi, et~al.]{guo2025deepseek}
Daya Guo, Dejian Yang, Haowei Zhang, Junxiao Song, Ruoyu Zhang, Runxin Xu, Qihao Zhu, Shirong Ma, Peiyi Wang, Xiao Bi, et~al.
\newblock Deepseek-r1: Incentivizing reasoning capability in llms via reinforcement learning.
\newblock \emph{arXiv preprint arXiv:2501.12948}, 2025.

\bibitem[Hong et~al.(2025)Hong, Yu, Gu, Wang, Gan, Tang, Cheng, Qi, Ji, Pan, et~al.]{hong2025glm}
Wenyi Hong, Wenmeng Yu, Xiaotao Gu, Guo Wang, Guobing Gan, Haomiao Tang, Jiale Cheng, Ji~Qi, Junhui Ji, Lihang Pan, et~al.
\newblock Glm-4.1 v-thinking: Towards versatile multimodal reasoning with scalable reinforcement learning.
\newblock \emph{arXiv e-prints}, pp.\  arXiv--2507, 2025.

\bibitem[Hurst et~al.(2024)Hurst, Lerer, Goucher, Perelman, Ramesh, Clark, Ostrow, Welihinda, Hayes, Radford, et~al.]{hurst2024gpt}
Aaron Hurst, Adam Lerer, Adam~P Goucher, Adam Perelman, Aditya Ramesh, Aidan Clark, AJ~Ostrow, Akila Welihinda, Alan Hayes, Alec Radford, et~al.
\newblock Gpt-4o system card.
\newblock \emph{arXiv preprint arXiv:2410.21276}, 2024.

\bibitem[{Kahatapitiya} et~al.(2024){Kahatapitiya}, {Ranasinghe}, {Park}, and {Ryoo}]{2024langrepo}
Kumara {Kahatapitiya}, Kanchana {Ranasinghe}, Jongwoo {Park}, and Michael~S. {Ryoo}.
\newblock {Language Repository for Long Video Understanding}.
\newblock \emph{arXiv e-prints}, art. arXiv:2403.14622, March 2024.
\newblock \doi{10.48550/arXiv.2403.14622}.

\bibitem[Kemp et~al.(2022)Kemp, Edsinger, Clever, and Matulevich]{kemp2022design}
Charles~C Kemp, Aaron Edsinger, Henry~M Clever, and Blaine Matulevich.
\newblock The design of stretch: A compact, lightweight mobile manipulator for indoor human environments.
\newblock In \emph{2022 International Conference on Robotics and Automation (ICRA)}, pp.\  3150--3157. IEEE, 2022.

\bibitem[{Khanna*} et~al.(2023){Khanna*}, {Mao*}, Jiang, Haresh, Shacklett, Batra, Clegg, Undersander, Chang, and Savva]{hssd}
Mukul {Khanna*}, Yongsen {Mao*}, Hanxiao Jiang, Sanjay Haresh, Brennan Shacklett, Dhruv Batra, Alexander Clegg, Eric Undersander, Angel~X. Chang, and Manolis Savva.
\newblock {Habitat Synthetic Scenes Dataset (HSSD-200): An Analysis of 3D Scene Scale and Realism Tradeoffs for ObjectGoal Navigation}.
\newblock \emph{arXiv preprint}, 2023.

\bibitem[Khanna* et~al.(2024)Khanna*, Ramrakhya*, Chhablani, Yenamandra, Gervet, Chang, Kira, Chaplot, Batra, and Mottaghi]{khanna2024goatbench}
Mukul Khanna*, Ram Ramrakhya*, Gunjan Chhablani, Sriram Yenamandra, Theophile Gervet, Matthew Chang, Zsolt Kira, Devendra~Singh Chaplot, Dhruv Batra, and Roozbeh Mottaghi.
\newblock Goat-bench: A benchmark for multi-modal lifelong navigation, 2024.

\bibitem[Kwon et~al.(2023)Kwon, Li, Zhuang, Sheng, Zheng, Yu, Gonzalez, Zhang, and Stoica]{kwon2023efficient}
Woosuk Kwon, Zhuohan Li, Siyuan Zhuang, Ying Sheng, Lianmin Zheng, Cody~Hao Yu, Joseph~E. Gonzalez, Hao Zhang, and Ion Stoica.
\newblock Efficient memory management for large language model serving with pagedattention.
\newblock In \emph{Proceedings of the ACM SIGOPS 29th Symposium on Operating Systems Principles}, 2023.

\bibitem[Li et~al.(2023)Li, Wang, Wang, Ge, Ge, and Shan]{li2023seed}
Bohao Li, Rui Wang, Guangzhi Wang, Yuying Ge, Yixiao Ge, and Ying Shan.
\newblock Seed-bench: Benchmarking multimodal llms with generative comprehension.
\newblock \emph{arXiv preprint arXiv:2307.16125}, 2023.

\bibitem[{Li} et~al.(2023){Li}, {Wang}, {He}, {Li}, {Wang}, {Liu}, {Wang}, {Xu}, {Chen}, {Luo}, {Wang}, and {Qiao}]{mvbench}
Kunchang {Li}, Yali {Wang}, Yinan {He}, Yizhuo {Li}, Yi~{Wang}, Yi~{Liu}, Zun {Wang}, Jilan {Xu}, Guo {Chen}, Ping {Luo}, Limin {Wang}, and Yu~{Qiao}.
\newblock {MVBench: A Comprehensive Multi-modal Video Understanding Benchmark}.
\newblock \emph{arXiv e-prints}, art. arXiv:2311.17005, November 2023.
\newblock \doi{10.48550/arXiv.2311.17005}.

\bibitem[Ma et~al.(2023)Ma, Yong, Zheng, Li, Liang, Zhu, and Huang]{ma2022sqa3d}
Xiaojian Ma, Silong Yong, Zilong Zheng, Qing Li, Yitao Liang, Song-Chun Zhu, and Siyuan Huang.
\newblock Sqa3d: Situated question answering in 3d scenes.
\newblock In \emph{International Conference on Learning Representations}, 2023.
\newblock URL \url{https://openreview.net/forum?id=IDJx97BC38}.

\bibitem[Ma et~al.(2024)Ma, Zang, Chen, Chen, Jiao, Li, Lu, Liu, Ma, Dong, Zhang, Pan, Jiang, Wang, Cao, and Sun]{longdocbench}
Yubo Ma, Yuhang Zang, Liangyu Chen, Meiqi Chen, Yizhu Jiao, Xinze Li, Xinyuan Lu, Ziyu Liu, Yan Ma, Xiaoyi Dong, Pan Zhang, Liangming Pan, Yu-Gang Jiang, Jiaqi Wang, Yixin Cao, and Aixin Sun.
\newblock Mmlongbench-doc: Benchmarking long-context document understanding with visualizations, 2024.
\newblock URL \url{https://arxiv.org/abs/2407.01523}.

\bibitem[Majumdar et~al.(2023)Majumdar, Yadav, Arnaud, Ma, Chen, Silwal, Jain, Berges, Wu, Vakil, et~al.]{majumdar2023we}
Arjun Majumdar, Karmesh Yadav, Sergio Arnaud, Jason Ma, Claire Chen, Sneha Silwal, Aryan Jain, Vincent-Pierre Berges, Tingfan Wu, Jay Vakil, et~al.
\newblock Where are we in the search for an artificial visual cortex for embodied intelligence?
\newblock \emph{Advances in Neural Information Processing Systems}, 36:\penalty0 655--677, 2023.

\bibitem[Majumdar et~al.(2024)Majumdar, Ajay, Zhang, Putta, Yenamandra, Henaff, Silwal, Mcvay, Maksymets, Arnaud, Yadav, Li, Newman, Sharma, Berges, Zhang, Agrawal, Bisk, Batra, Kalakrishnan, Meier, Paxton, Sax, and Rajeswaran]{OpenEQA2023}
Arjun Majumdar, Anurag Ajay, Xiaohan Zhang, Pranav Putta, Sriram Yenamandra, Mikael Henaff, Sneha Silwal, Paul Mcvay, Oleksandr Maksymets, Sergio Arnaud, Karmesh Yadav, Qiyang Li, Ben Newman, Mohit Sharma, Vincent Berges, Shiqi Zhang, Pulkit Agrawal, Yonatan Bisk, Dhruv Batra, Mrinal Kalakrishnan, Franziska Meier, Chris Paxton, Sasha Sax, and Aravind Rajeswaran.
\newblock Openeqa: Embodied question answering in the era of foundation models.
\newblock In \emph{Conference on Computer Vision and Pattern Recognition (CVPR)}, 2024.

\bibitem[{Mangalam} et~al.(2023){Mangalam}, {Akshulakov}, and {Malik}]{egoschema}
Karttikeya {Mangalam}, Raiymbek {Akshulakov}, and Jitendra {Malik}.
\newblock {EgoSchema: A Diagnostic Benchmark for Very Long-form Video Language Understanding}.
\newblock \emph{arXiv e-prints}, art. arXiv:2308.09126, August 2023.
\newblock \doi{10.48550/arXiv.2308.09126}.

\bibitem[{Min} et~al.(2024){Min}, {Puig}, {Singh Chaplot}, {Yang}, {Rai}, {Parashar}, {Salakhutdinov}, {Bisk}, and {Mottaghi}]{2024sif}
So~Yeon {Min}, Xavi {Puig}, Devendra {Singh Chaplot}, Tsung-Yen {Yang}, Akshara {Rai}, Priyam {Parashar}, Ruslan {Salakhutdinov}, Yonatan {Bisk}, and Roozbeh {Mottaghi}.
\newblock {Situated Instruction Following}.
\newblock \emph{arXiv e-prints}, art. arXiv:2407.12061, July 2024.
\newblock \doi{10.48550/arXiv.2407.12061}.

\bibitem[Ni et~al.(2023)Ni, Ma, Eysenbach, and Bacon]{ni2023when}
Tianwei Ni, Michel Ma, Benjamin Eysenbach, and Pierre-Luc Bacon.
\newblock When do transformers shine in {RL}? decoupling memory from credit assignment.
\newblock In \emph{Thirty-seventh Conference on Neural Information Processing Systems}, 2023.
\newblock URL \url{https://openreview.net/forum?id=APGXBNkt6h}.

\bibitem[Pa{\v{s}}ukonis et~al.(2023)Pa{\v{s}}ukonis, Lillicrap, and Hafner]{pasukonis2023evaluating}
Jurgis Pa{\v{s}}ukonis, Timothy~P Lillicrap, and Danijar Hafner.
\newblock Evaluating long-term memory in 3d mazes.
\newblock In \emph{The Eleventh International Conference on Learning Representations}, 2023.
\newblock URL \url{https://openreview.net/forum?id=yHLvIlE9RGN}.

\bibitem[Pleines et~al.(2023)Pleines, Pallasch, Zimmer, and Preuss]{pleines2023memory}
Marco Pleines, Matthias Pallasch, Frank Zimmer, and Mike Preuss.
\newblock Memory gym: Partially observable challenges to memory-based agents.
\newblock In \emph{International Conference on Learning Representations}, 2023.
\newblock URL \url{https://openreview.net/forum?id=jHc8dCx6DDr}.

\bibitem[Ramakrishnan et~al.(2024)Ramakrishnan, Wijmans, Kraehenbuehl, and Koltun]{ramakrishnan2024does}
Santhosh~Kumar Ramakrishnan, Erik Wijmans, Philipp Kraehenbuehl, and Vladlen Koltun.
\newblock Does spatial cognition emerge in frontier models?
\newblock \emph{arXiv preprint arXiv:2410.06468}, 2024.

\bibitem[Rawal et~al.(2024)Rawal, Saifullah, Basri, Jacobs, Somepalli, and Goldstein]{rawal2024cinepile}
Ruchit Rawal, Khalid Saifullah, Ronen Basri, David Jacobs, Gowthami Somepalli, and Tom Goldstein.
\newblock Cinepile: A long video question answering dataset and benchmark.
\newblock \emph{arXiv preprint arXiv:2405.08813}, 2024.

\bibitem[Ren et~al.(2024)Ren, Clark, Dixit, Itkina, Majumdar, and Sadigh]{exploreeqa2024}
Allen~Z. Ren, Jaden Clark, Anushri Dixit, Masha Itkina, Anirudha Majumdar, and Dorsa Sadigh.
\newblock Explore until confident: Efficient exploration for embodied question answering.
\newblock In \emph{arXiv preprint arXiv:2403.15941}, 2024.

\bibitem[{Sashank Dorbala} et~al.(2024){Sashank Dorbala}, {Goyal}, {Piramuthu}, {Johnston}, {Ghanadhan}, and {Manocha}]{seqa}
Vishnu {Sashank Dorbala}, Prasoon {Goyal}, Robinson {Piramuthu}, Michael {Johnston}, Reza {Ghanadhan}, and Dinesh {Manocha}.
\newblock {S-EQA: Tackling Situational Queries in Embodied Question Answering}.
\newblock \emph{arXiv e-prints}, art. arXiv:2405.04732, May 2024.
\newblock \doi{10.48550/arXiv.2405.04732}.

\bibitem[Savva et~al.(2019)Savva, Kadian, Maksymets, Zhao, Wijmans, Jain, Straub, Liu, Koltun, Malik, Parikh, and Batra]{habitat19iccv}
Manolis Savva, Abhishek Kadian, Oleksandr Maksymets, Yili Zhao, Erik Wijmans, Bhavana Jain, Julian Straub, Jia Liu, Vladlen Koltun, Jitendra Malik, Devi Parikh, and Dhruv Batra.
\newblock Habitat: {A} {P}latform for {E}mbodied {AI} {R}esearch.
\newblock \emph{ICCV}, 2019.

\bibitem[Sethian(1996)]{fmm}
James~A Sethian.
\newblock A fast marching level set method for monotonically advancing fronts.
\newblock \emph{proceedings of the National Academy of Sciences}, 93\penalty0 (4):\penalty0 1591--1595, 1996.

\bibitem[Song et~al.(2024)Song, Chen, Chen, Yu, Wan, and Wang]{song2024milebench}
Dingjie Song, Shunian Chen, Guiming~Hardy Chen, Fei Yu, Xiang Wan, and Benyou Wang.
\newblock Milebench: Benchmarking mllms in long context.
\newblock \emph{arXiv preprint arXiv:2404.18532}, 2024.

\bibitem[Song et~al.(2023)Song, Chai, Wang, Zhang, Zhou, Wu, Guo, Ye, Lu, Hwang, et~al.]{song2023moviechat}
Enxin Song, Wenhao Chai, Guanhong Wang, Yucheng Zhang, Haoyang Zhou, Feiyang Wu, Xun Guo, Tian Ye, Yan Lu, Jenq-Neng Hwang, et~al.
\newblock Moviechat: From dense token to sparse memory for long video understanding.
\newblock \emph{arXiv preprint arXiv:2307.16449}, 2023.

\bibitem[Szot et~al.(2021)Szot, Clegg, Undersander, Wijmans, Zhao, Turner, Maestre, Mukadam, Chaplot, Maksymets, et~al.]{szot2021habitat}
Andrew Szot, Alexander Clegg, Eric Undersander, Erik Wijmans, Yili Zhao, John Turner, Noah Maestre, Mustafa Mukadam, Devendra~Singh Chaplot, Oleksandr Maksymets, et~al.
\newblock Habitat 2.0: Training home assistants to rearrange their habitat.
\newblock In \emph{NeurIPS}, 2021.

\bibitem[Szot et~al.(2024)Szot, Schwarzer, Agrawal, Mazoure, Metcalf, Talbott, Mackraz, Hjelm, and Toshev]{szot2024large}
Andrew Szot, Max Schwarzer, Harsh Agrawal, Bogdan Mazoure, Rin Metcalf, Walter Talbott, Natalie Mackraz, R~Devon Hjelm, and Alexander~T Toshev.
\newblock Large language models as generalizable policies for embodied tasks.
\newblock In \emph{The Twelfth International Conference on Learning Representations}, 2024.
\newblock URL \url{https://openreview.net/forum?id=u6imHU4Ebu}.

\bibitem[Team et~al.(2024)Team, Georgiev, Lei, Burnell, Bai, Gulati, Tanzer, Vincent, Pan, Wang, et~al.]{team2024gemini}
Gemini Team, Petko Georgiev, Ving~Ian Lei, Ryan Burnell, Libin Bai, Anmol Gulati, Garrett Tanzer, Damien Vincent, Zhufeng Pan, Shibo Wang, et~al.
\newblock Gemini 1.5: Unlocking multimodal understanding across millions of tokens of context.
\newblock \emph{arXiv preprint arXiv:2403.05530}, 2024.

\bibitem[Team et~al.(2025)Team, Kamath, Ferret, Pathak, Vieillard, Merhej, Perrin, Matejovicova, Ram{\'e}, Rivi{\`e}re, et~al.]{team2025gemma}
Gemma Team, Aishwarya Kamath, Johan Ferret, Shreya Pathak, Nino Vieillard, Ramona Merhej, Sarah Perrin, Tatiana Matejovicova, Alexandre Ram{\'e}, Morgane Rivi{\`e}re, et~al.
\newblock Gemma 3 technical report.
\newblock \emph{arXiv preprint arXiv:2503.19786}, 2025.

\bibitem[Wang et~al.(2024)Wang, Ning, Pan, Wu, Guo, Deng, Bao, Wang, and Zhang]{wang2024novelqa}
Cunxiang Wang, Ruoxi Ning, Boqi Pan, Tonghui Wu, Qipeng Guo, Cheng Deng, Guangsheng Bao, Qian Wang, and Yue Zhang.
\newblock Novelqa: A benchmark for long-range novel question answering, 2024.

\bibitem[Wani et~al.(2020)Wani, Patel, Jain, Chang, and Savva]{wani2020multion}
Saim Wani, Shivansh Patel, Unnat Jain, Angel Chang, and Manolis Savva.
\newblock Multion: Benchmarking semantic map memory using multi-object navigation.
\newblock \emph{NeurIPS}, 2020.

\bibitem[{Xie} et~al.(2024){Xie}, {Min}, {Zhang}, {Xu}, {Bajaj}, {Salakhutdinov}, {Johnson-Roberson}, and {Bisk}]{emb-rag}
Quanting {Xie}, So~Yeon {Min}, Tianyi {Zhang}, Kedi {Xu}, Aarav {Bajaj}, Ruslan {Salakhutdinov}, Matthew {Johnson-Roberson}, and Yonatan {Bisk}.
\newblock {Embodied-RAG: General Non-parametric Embodied Memory for Retrieval and Generation}.
\newblock \emph{arXiv e-prints}, art. arXiv:2409.18313, September 2024.
\newblock \doi{10.48550/arXiv.2409.18313}.

\bibitem[Xiong et~al.(2025)Xiong, Yang, Yu, Zhuge, Zhang, Zhu, and Lu]{streamchat}
Haomiao Xiong, Zongxin Yang, Jiazuo Yu, Yunzhi Zhuge, Lu~Zhang, Jiawen Zhu, and Huchuan Lu.
\newblock Streaming video understanding and multi-round interaction with memory-enhanced knowledge, 2025.
\newblock URL \url{https://arxiv.org/abs/2501.13468}.

\bibitem[Xu et~al.(2024)Xu, Chiang, Fu, Jacob, Zhang, Lee, Yu, Schenck, Rendleman, Shah, Xia, Hsu, Hoech, Florence, Kirmani, Singh, Sindhwani, Parada, Finn, Xu, Levine, and Tan]{xu2024mobility}
Zhuo Xu, Hao-Tien~Lewis Chiang, Zipeng Fu, Mithun~George Jacob, Tingnan Zhang, Tsang-Wei~Edward Lee, Wenhao Yu, Connor Schenck, David Rendleman, Dhruv Shah, Fei Xia, Jasmine Hsu, Jonathan Hoech, Pete Florence, Sean Kirmani, Sumeet Singh, Vikas Sindhwani, Carolina Parada, Chelsea Finn, Peng Xu, Sergey Levine, and Jie Tan.
\newblock Mobility {VLA}: Multimodal instruction navigation with long-context {VLM}s and topological graphs.
\newblock In \emph{8th Annual Conference on Robot Learning}, 2024.
\newblock URL \url{https://openreview.net/forum?id=JScswMfEQ0}.

\bibitem[Yadav et~al.(2023)Yadav, Majumdar, Ramrakhya, Yokoyama, Baevski, Kira, Maksymets, and Batra]{yadav2023ovrlv2}
Karmesh Yadav, Arjun Majumdar, Ram Ramrakhya, Naoki Yokoyama, Alexei Baevski, Zsolt Kira, Oleksandr Maksymets, and Dhruv Batra.
\newblock Ovrl-v2: A simple state-of-art baseline for imagenav and objectnav.
\newblock \emph{\url{https://arxiv.org/abs/2303.07798}}, 2023.

\bibitem[Yang et~al.(2024)Yang, Yang, Gupta, Han, Fei-Fei, and Xie]{yang2024thinking}
Jihan Yang, Shusheng Yang, Anjali~W Gupta, Rilyn Han, Li~Fei-Fei, and Saining Xie.
\newblock Thinking in space: How multimodal large language models see, remember, and recall spaces.
\newblock \emph{arXiv preprint arXiv:2412.14171}, 2024.

\bibitem[Yenamandra et~al.(2023)Yenamandra, Ramachandran, Yadav, Wang, Khanna, Gervet, Yang, Jain, Clegg, Turner, et~al.]{yenamandra2023homerobot}
Sriram Yenamandra, Arun Ramachandran, Karmesh Yadav, Austin Wang, Mukul Khanna, Theophile Gervet, Tsung-Yen Yang, Vidhi Jain, Alexander~William Clegg, John Turner, et~al.
\newblock Homerobot: Open-vocabulary mobile manipulation.
\newblock \emph{arXiv preprint arXiv:2306.11565}, 2023.

\bibitem[Zhang et~al.(2023)Zhang, Lu, Islam, Wang, Yu, Bansal, and Bertasius]{zhang2023simplellovi}
Ce~Zhang, Taixi Lu, Md~Mohaiminul Islam, Ziyang Wang, Shoubin Yu, Mohit Bansal, and Gedas Bertasius.
\newblock A simple llm framework for long-range video question-answering, 2023.

\bibitem[Zhang et~al.(2024)Zhang, Zhang, Li, Zeng, Yang, Zhang, Wang, Tan, Li, and Liu]{zhang2024longva}
Peiyuan Zhang, Kaichen Zhang, Bo~Li, Guangtao Zeng, Jingkang Yang, Yuanhan Zhang, Ziyue Wang, Haoran Tan, Chunyuan Li, and Ziwei Liu.
\newblock Long context transfer from language to vision.
\newblock \emph{arXiv preprint arXiv:2406.16852}, 2024.
\newblock URL \url{https://arxiv.org/abs/2406.16852}.

\bibitem[Zhou et~al.(2024)Zhou, Shu, Zhao, Wu, Xiao, Yang, Xiong, Zhang, Huang, and Liu]{mvlu}
Junjie Zhou, Yan Shu, Bo~Zhao, Boya Wu, Shitao Xiao, Xi~Yang, Yongping Xiong, Bo~Zhang, Tiejun Huang, and Zheng Liu.
\newblock Mlvu: A comprehensive benchmark for multi-task long video understanding.
\newblock \emph{arXiv preprint arXiv:2406.04264}, 2024.

\bibitem[Zhu et~al.(2023)Zhu, Kapoor, Min, Han, Li, Geng, Neubig, Bisk, Kembhavi, and Weihs]{Zhu_2023_CVPR}
Hao Zhu, Raghav Kapoor, So~Yeon Min, Winson Han, Jiatai Li, Kaiwen Geng, Graham Neubig, Yonatan Bisk, Aniruddha Kembhavi, and Luca Weihs.
\newblock Excalibur: Encouraging and evaluating embodied exploration.
\newblock In \emph{Proceedings of the IEEE/CVF Conference on Computer Vision and Pattern Recognition (CVPR)}, pp.\  14931--14942, June 2023.

\end{thebibliography}
\bibliographystyle{iclr2026_conference}

\newpage
\appendix
{
   \onecolumn
    \centering
    \Large
    \vspace{0.5em}\textbf{Supplementary Material} \\
    \vspace{0.5em}
}

\section{Limitations}
While \ourbench provides a rigorous evaluation framework for long-horizon memory in embodied AI, it has certain limitations that highlight areas for future improvement.

First, our most competitive baseline, which integrates a VLM with a navigation policy, is constrained by its low-level controller. Current memory-guided navigation policies struggle to take the shortest paths in large environments after reasoning over long interaction histories. As a result, models perform poorly on spatial awareness metrics, making it difficult to assess their true memory capabilities beyond basic path efficiency. Future work should explore improved memory-driven navigation policies to better leverage long-term spatial reasoning.

Second, in the experience collection phase, our scripted oracle policy for object interactions (Magic Grasp) executes pick-and-place actions in an unnatural way—objects are abruptly transferred from the receptacle surface to the agent's gripper without smooth motion. This may confuse VLMs, as the sudden change is visually subtle and may be overlooked if the model does not attend to the specific frames where it occurs. Developing more naturalistic interaction models could improve the realism and interpretability of memory traces.

Third, while we introduce distractor objects to discourage guessing, we do not explicitly control how prominently these objects appear in the experience collection phase. Currently, objects the agent interacts with are often viewed more frequently and closely compared to distractors, which tend to be featured less prominently in the agent’s field of view. Although some tasks mitigate this by explicitly referencing non-interacted objects, future iterations of the benchmark may need stronger distractions by ensuring more balanced exposure between distractors and interacted objects.

Finally, while the experience collection phase involves both navigation and manipulation, the current interaction phase tasks are all navigation-based, as memory reasoning is primarily required for locating objects rather than executing pick-and-place actions, which are Markovian once the goal is known. However, our benchmark is easily extensible to include manipulation-based memory tasks, which would further enrich the evaluation of long-term reasoning in embodied AI.

Despite these limitations, \ourbench establishes a strong foundation for studying long-horizon memory, offering extensible challenges that will evolve alongside advances in vision-language models and embodied learning.

\section{Benchmark Details}
\label{sec:dataset_creation}
\subsection{\ourbench Episode Creation}
In this section, we give details about our episode creation pipeline that allows us to procedurally generate data for our benchmark.

\begin{figure*}[ht]
    \centering
    \begin{subfigure}{\linewidth}
        \centering
        \includegraphics[width=\linewidth]{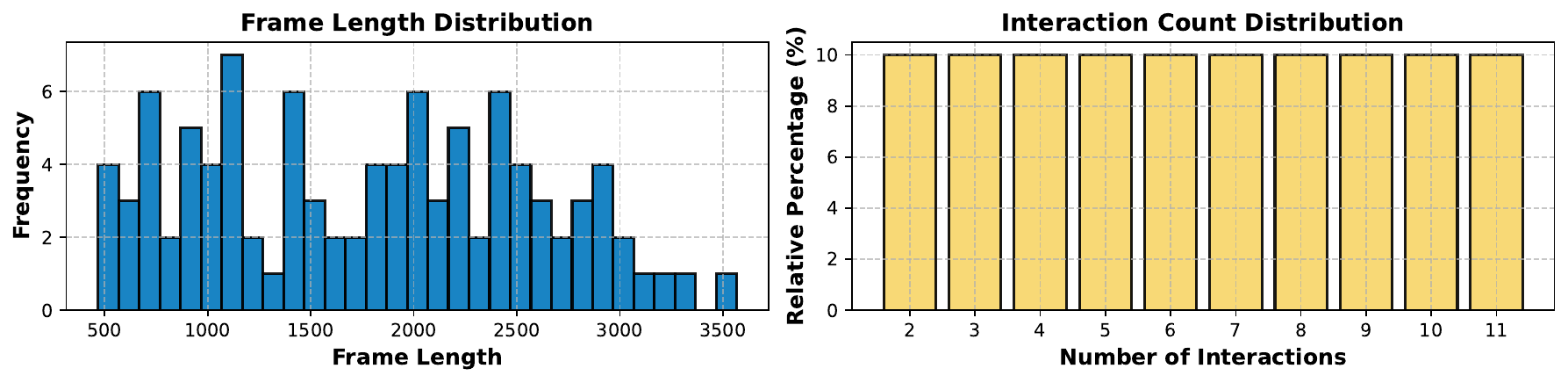}
        \vspace{-15pt}
        \caption{Distributions of frame counts and interaction counts across episodes.}
        \label{fig:interaction_dist}
    \end{subfigure}
    
    \begin{subfigure}{0.49\linewidth}
        \centering
        \includegraphics[width=\linewidth]{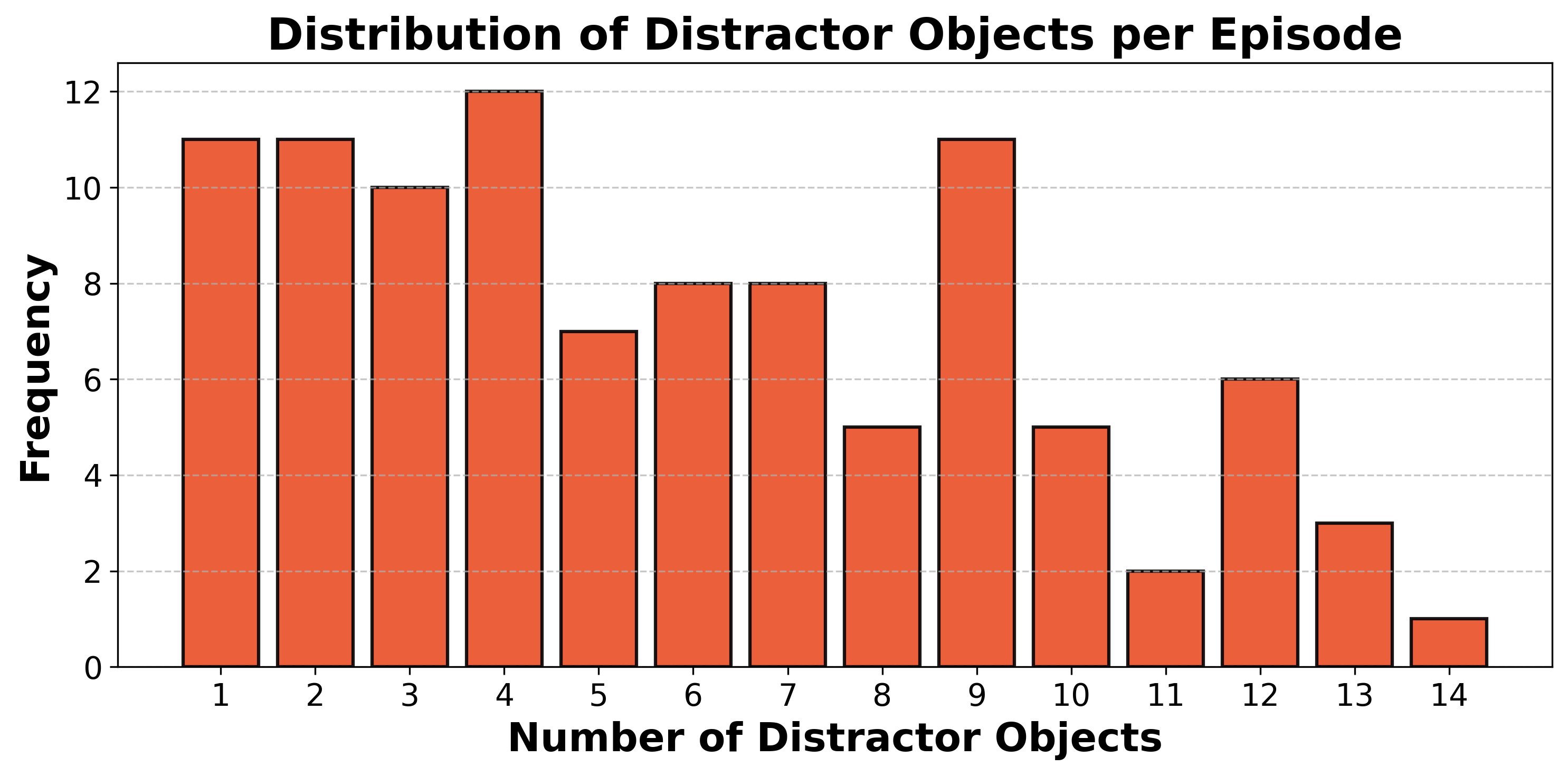}
        \caption{Number of objects per episode.}
        \label{fig:distrator_objects_per_episode}
    \end{subfigure}
    \hfill
    \begin{subfigure}{0.49\linewidth}
        \centering
        \includegraphics[width=\linewidth]{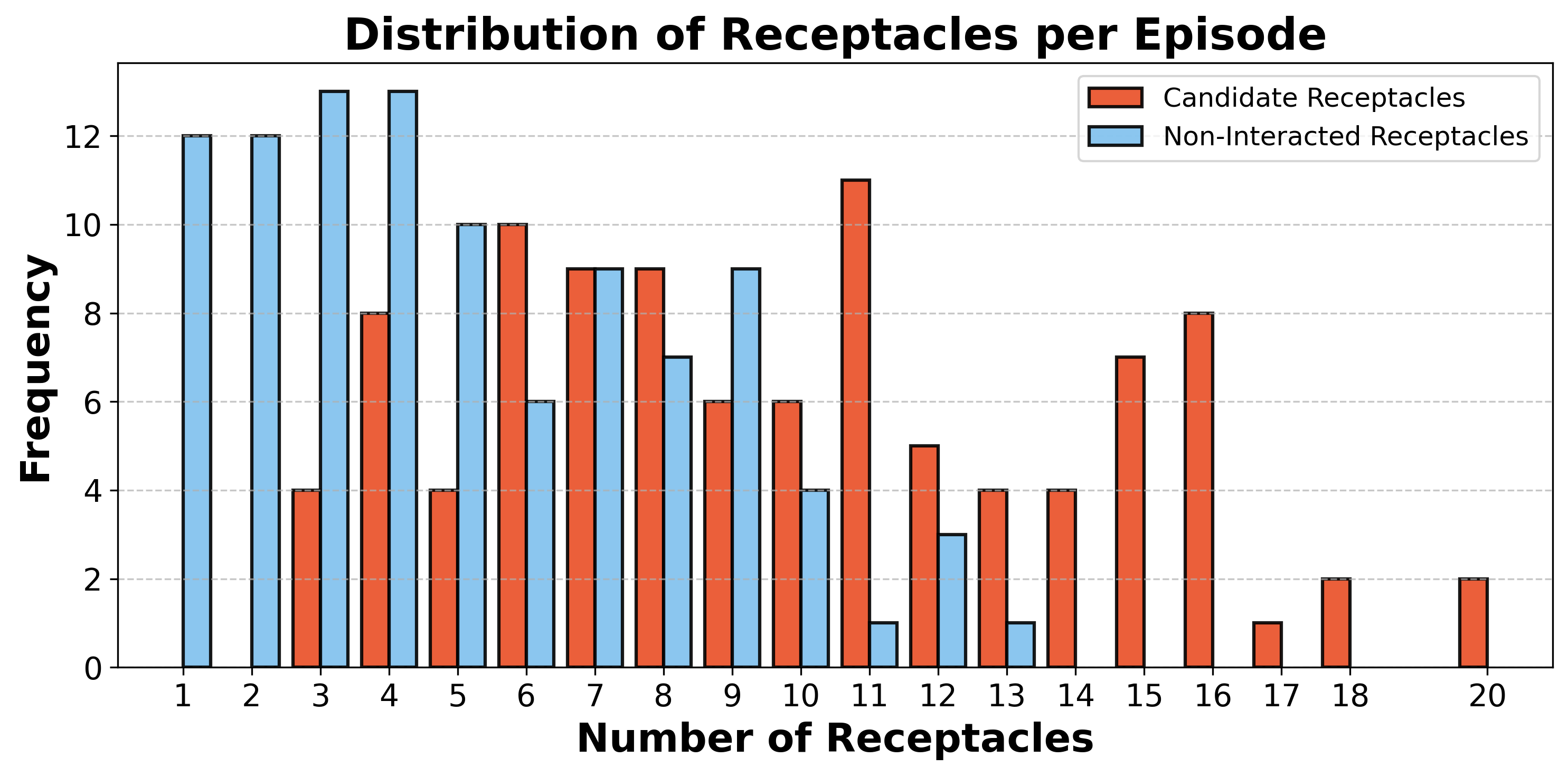}
        \vspace{-10pt}
        \caption{Number of receptacles per episode.}
        \label{fig:receptacles_per_episode}
    \end{subfigure}
    \vspace{-10pt}
    \caption{Summary statistics of episode data.}
    \vspace{-10pt}
    \label{fig:merged_distributions}
\end{figure*}

\textbf{Object-Receptacle Pairing.} The first step in the episode creation involves generating valid spawning positions for various object-receptacle pairs by sampling entities from the respective split in consideration (\texttt{train/val}). The objective is to identify \textit{candidate objects} that can be picked up from a \textit{start receptacle} and placed stably on a \textit{goal receptacle}. We use the pipeline proposed in~\citet{yenamandra2023homerobot} to procedurally generate various object placements on receptacle surfaces by running physics checks for stable placements. To make the generation procedure easier, we assume that each object is picked and placed only once during an episode. Additionally, we constrain the \textit{start receptacle} category to be distinct from the \textit{goal receptacle} category. We also ensure that for each \textit{candidate object} category, a corresponding \textit{non-interacted object} of the same category is spawned to serve as distractor entities during the interaction phase. ~\Cref{fig:distrator_objects_per_episode,fig:receptacles_per_episode} provide a distribution of distractor objects and receptacles in the val episodes. The number of sampled object-receptacle pairings is equal to the number of rearrangements in the particular episode and can be configured through an argument. In essence, this controls the distribution of video frame lengths in the \ourbench tasks(see ~\Cref{fig:interaction_dist}).

\textbf{Object Placement Sequences.} Once we have identified the set of \textit{candidate objects} and the corresponding \textit{goal receptacles}, we require an oracle policy through which we can ``naturally" drop objects on the receptacle surfaces. In practice, one can achieve this by training a ``place" skill through end-to-end reinforcement learning or use a heuristic agent that can place objects on receptacles but both frameworks do not guarantee perfect $100\%$ success which is crucial for solvability in \ourbench tasks. To address this, we construct an offline pipeline that produces the set of low-level ``oracle" joint actions that lead to stably placing the object on the goal receptacle. We spawn the agent (with the object snapped in its end-effector) at the associated viewpoints of \textit{goal receptacles} on the navigation mesh. From each viewpoint, we try to execute a heuristic, state-machine based placing policy which performs a set of predefined joint effector actions after identifying a ``clutter-free" region in the receptacle pointcloud (using an onboard depth sensor). The output of this routine produces a set of low level joint actions and the corresponding receptacle viewpoint for each \textit{goal receptacle} that was identified in the previous object-receptacle pairing stage. We reject episodes where the heuristic-based place policy cannot stably place the object from \textit{any} of the associated goal receptacle viewpoints.

\textbf{Task Creation.} Once we have the final filtered set of episodes, we populate task instructions for each episode using \texttt{yaml}-based instruction templates as used in ~\citet{szot2024large}. 
We employ the instruction templates to \textit{procedurally} populate various placeholders such as object names, attributes and appearance orders that create specific task instructions grounded in the entities (objects and receptacles) present in the episode in an automated manner. 
We present the diverse instructions templates we use for each task category in Table \ref{tab:all_tasks}. 
In total, we create $82174/5876$ unique task instructions within the \texttt{train/val} split respectively.

\vspace{-8pt}

\subsection{Experience Collection Phase}
\label{appendix:experience_collection}

We now outline the details of the experience collection phase grounded in the episodes created by the pipeline outlined in the previous section. The experience collection phase is designed to collect a ``clean" and ``natural" video sequence of an agent interacting in an home environment while interacting with various entities. We focus on \textit{object rearrangements} as the core interaction routine followed by the embodied agent. We construct an \texttt{OracleAgent} agent that uses privileged information from the simulator instance to efficiently generate such interaction routines. We now detail the \texttt{OracleAgent} policy to generate such video sequences for a particular episode.

\textbf{Nav-to-Pick.} The agent begins by being spawned at a random location in the scene associated with a particular episode. We then query a \texttt{ShortestPath} function that leverages privileged simulator information to navigate to the object that has to be picked up. The \texttt{OracleAgent} follows the shortest path to the object and collects images as it navigates in the environment.

\textbf{Pick.} Once the \texttt{OracleAgent} reaches the object location, it executes a hardcoded pick policy sequence which results in the object being magically grasped by the \texttt{OracleAgent}. The pick policy uses the onboard depth sensor to orient itself with the object instance placed on the receptacle infront of the agent. Once the object is in clear view (using the ground truth instance mask from the simulator), the and effector extends outwards to perform a \texttt{magic\_grasp} action. In cases when the object is not visible, we re-query the \texttt{ShortestPath} function to navigate to an alternate viewpoint associated with the receptacle that would lead to the object being clearly visible.

\textbf{Nav-to-Place.} After grasping the object, the \texttt{OracleAgent} navigates to the corresponding \textit{goal receptacle} that was sampled during the episode creation phase. Specifically, the agent navigates to the specific viewpoint associated with the \textit{goal receptacle} from which the pre-recorded placement action sequence is to be executed.

\textbf{Place.} Once the agent reaches the specific viewpoint, we re-play the end-effector action sequence that was found to lead to a stable object placement during episode creation. This marks the end of a single rearrangement interaction routine and the agent transitions to rearrange the next \textit{candidate object} in the episode until no more objects are left.

\textbf{Final Navigation.} At the end of all interaction routines within an episode, the agent navigates to a predefined location to ensure that it does not remain in close proximity to the final interacted object or the final \textit{goal\_receptacle}. This ensures that tasks are not solvable by selecting only the last few frames in the video sequence. We ensure that the agent navigates to a distance of $3$ m away from the final interaction location.

The above \texttt{OracleAgent} routine enables us to procedurally create videos of the agent rearranging objects in the environment. Additionally, as the agent navigates between the pick and place location, it collects additional frames detailing the scene environment which can be useful (or serve as distractor frames) when attempting to solve \ourbench tasks. We present an example experience collection trajectory and associated ``magic" pick-place sequence in \Cref{fig:qualitative_traj,fig:oracle_pick_place}. %

\subsection{Evaluations}
\label{app:evaluations}
In this section, we define the primary evaluation metrics employed to benchmark the performance of various baselines on the \ourbench tasks. Following~\citet{batra2020objectnav}, we focus on metrics that systematically quantify both the success and efficiency of the agent in solving specified instructions. Success focuses on accurately selecting frames from the interaction video that lead to task completion, while efficiency evaluates if the agent selects and navigates to the optimal frame that solves the task using the fewest low-level control actions. Notably, optimizing efficiency poses a significant challenge for current methods~\citep{ramakrishnan2024does,yang2024thinking}. 

\textbf{High-Level Policy Success Rate} (\texttt{HL-SR}). This metric measures the percentage of episodes in which the high-level agent correctly predicts frame indices required to solve the specified task. We define an episode as successful if the frame(s) predicted by the VLM satisfy the following criteria:

\begin{enumerate}[leftmargin=2em, labelindent=0.5em, labelsep=0.5em]
    \item \textit{Distance-to-Goal}: The agent is within a specified distance of the target entity upon reaching the predicted frame. We use thresholds of 2.0 m for objects and 0.1 m for receptacles. The larger threshold for objects accounts for the possibility of objects being positioned on elongated receptacles (e.g., couches), potentially affecting reachability.
    \item \textit{Angle-to-Goal}: The agent is oriented within a specified angular distance toward the target entity's center. We set angle thresholds of $45^\circ$ for objects and $90^\circ$ for receptacles due to their typically larger dimensions.
    \item \textit{Semantic Coverage}: The semantic mask of the target entity covers at least $0.1\%$ of the pixels in the frame.
    \item \textit{Room Region Check}: For room visitation tasks, we verify if the agent is located within the designated target region.
\end{enumerate}

These thresholds and criteria were qualitatively validated through manual inspection to ensure their appropriateness; During evaluation, these criteria are verified by teleporting the robot to the pose associated with the frame(s) selected by the high-level agent.

\textbf{Low-Level Policy Success Rate} (\texttt{LL-SR}). This metric measures the percentage of episodes where the low-level navigation agent successfully navigates to the target entity. The low-level agent is activated only if the high-level agent correctly predicts the frame indices, meaning \texttt{LL-SR} is evaluated conditionally upon \texttt{HL-SR} being true for each episode. The success criteria match those outlined for \texttt{HL-SR}.

\textbf{Success-weighted-by-Path-Length}. This metric evaluates task completion efficiency, computed per task episode as follows~\citep{batra2020objectnav}:
\begin{equation*}
    \text{SPL}_i = S_i \cdot \frac{l_i}{\max(p_i, l_i)}
\end{equation*}

where $S_i$ denotes \texttt{HL-SR/LL-SR} for computing the \texttt{HL-SPL/LL-SPL} respectively, $l_i$ is the length of the shortest possible path to the closest successful goal frame (or shortest path to revisit all subgoals), and $p_i$ is the actual path length traveled by the agent.

Additionally, we introduce two relaxed success metrics to analyze baseline failure modes:
\vspace{-5pt}
\begin{enumerate}[leftmargin=2em, labelindent=0.5em, labelsep=0.5em]
    \item \textbf{Distance-to-Goal-Only Success Rate} (\texttt{DTG-SR}). Computed identically to \texttt{HL-SR} but without the semantic coverage requirement. This metric helps quantify instances where the agent selects frames close to the goal without necessarily capturing it visually.
    \item \textbf{Semantic Coverage Success Rate} (\texttt{SC-SR}). Computed identically to \texttt{HL-SR} but without the distance-to-goal requirement. This metric evaluates instances where the agent correctly selects frames showing the target entity but fails to meet proximity criteria.
\end{enumerate}

\subsection{Subsampled Videos Solvability Analysis}
\label{app:task_solvability}

\begin{wrapfigure}{r}{0.4\linewidth} %
    \centering
    \includegraphics[width=\linewidth]{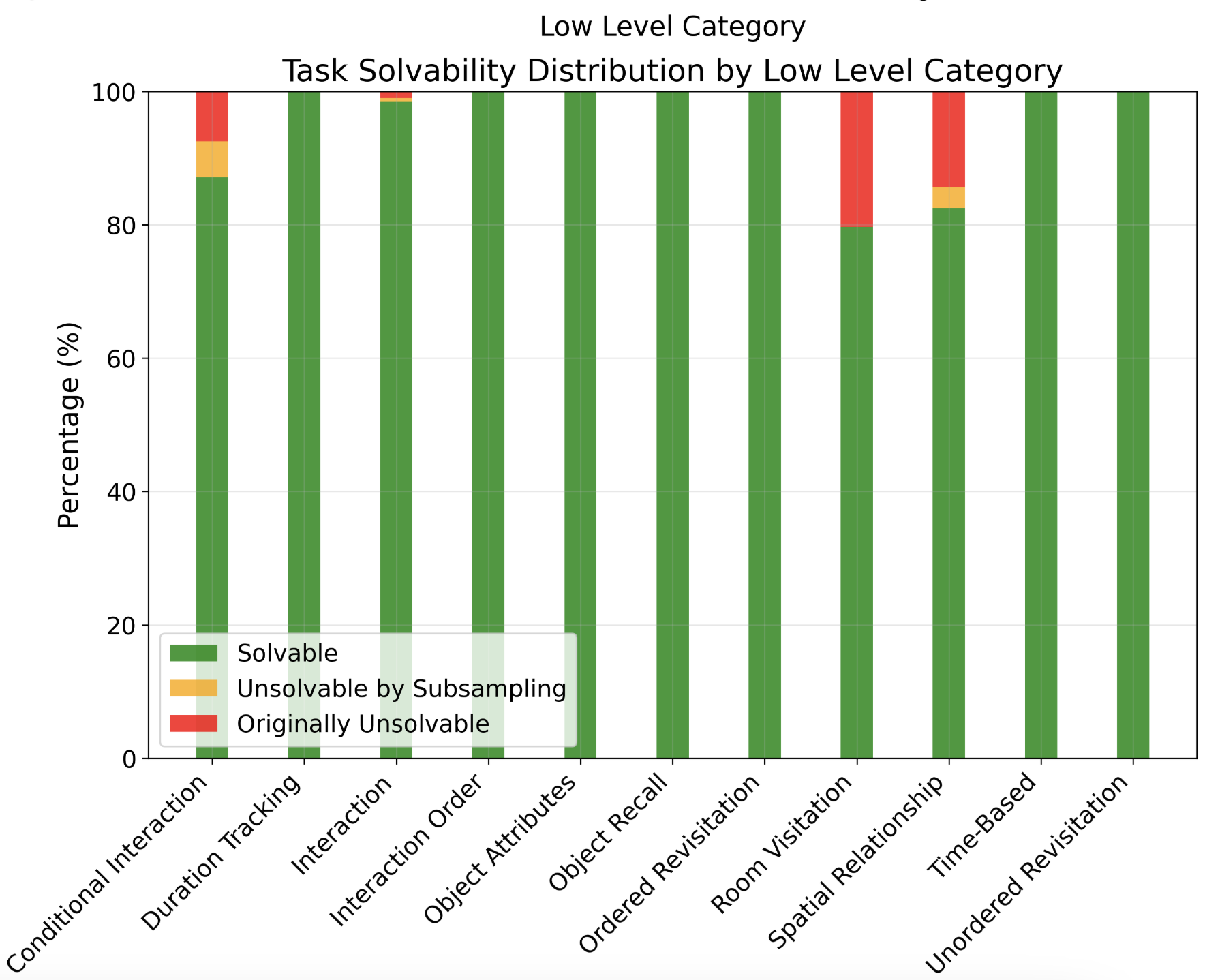}
    \caption{Task Solvability Ratio at the $96$ frame subsampling level.}
    \vspace{-10pt}
    \label{fig:task_solvability_96}
\end{wrapfigure}

In this section, we analyze the solvability of the tasks proposed in \ourbench suite when all videos are subsampled to enable full-finetuning of VLMs (see~\cref{appendix:sft_baseline} for details).
Specifically, we conduct a solvability analysis at the $96$-frame video subsampling level. 
Full finetuning of the VLM on these videos enables highest task performance of $\approx 52\%$ as shown in ~\cref{fig:model_checkpoint_comparison}.

From ~\cref{fig:task_solvability_96} we observe that after subsampling the videos to $96$ frames, there is a $\approx1\%$ drop in solvability across the various low-level task categories. 
The largest drop is observed in the \textit{Conditional Interaction} tasks where a specific object or receptacle is to be revisited based on the original interaction sequence. 
We want to highlight that this solvability calculation is solely based on whether a \textit{solution frame} (from multiple possible frames in the full length video) is left over in the subsampled video. 
Such frames are treated as a solution based on whether they satisfy the PDDL goal conditions (see ~\cref{sec:implementation_details}) corresponding to the task.
But such frames are extremely difficult to select for goal completion for a VLM since the subsampling routine causes a large drop in the relevant context that is needed to make sense of the solution frame.
For instance, consider the task \textit{``Navigate to the object which you interacted with which is the farthest from your current
location"}: it is possible that a frame with the target object is left over in the subsampled video. 
But selecting this solution frame is nearly impossible since the subsampling to $96$ frames eliminates all the spatiotemporal context that is necessary to solve this task.

From ~\cref{fig:task_solvability_96} we also observe that $3.85\%$ of the tasks are \textit{Originally Unsolvable} which correspond to tasks focused on distractor entities such as \textit{``Navigate to a soft toy that you did not rearrange yesterday"} or \textit{``Navigate to a room that you did not visit yesterday"}.

\subsection{Chain-of-Thought Trace Generation}
\label{app:cot_generation}

\begin{figure*}[!htb]
  \vspace{-10pt}
  \centering
  \includegraphics[width=\linewidth]{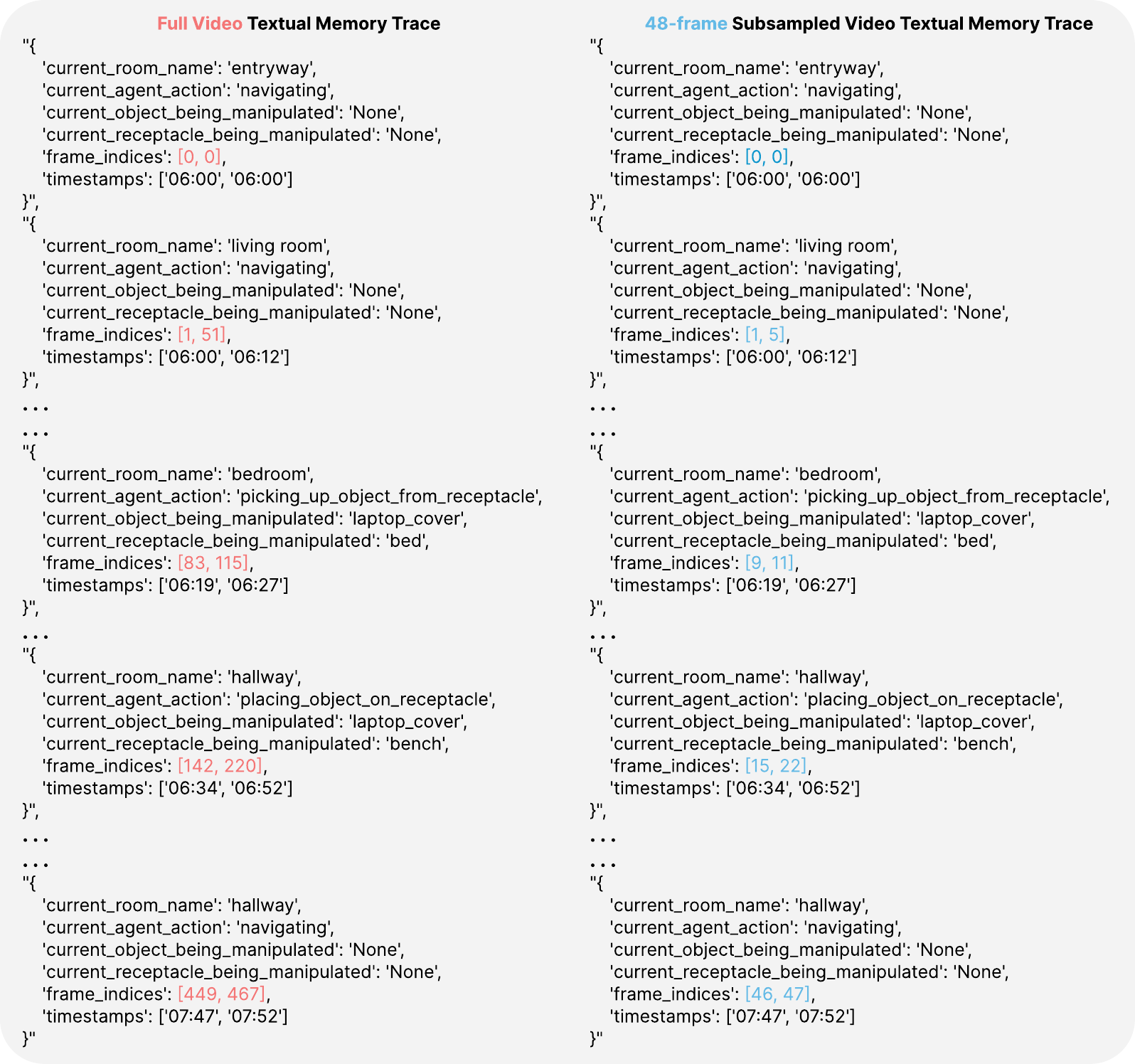}
  \vspace{-10pt}
  \caption{Textual Memory Trace corresponding to agent interaction during \textit{Experience Collection}. The memory trace is subsampled to ensure alignment with the 48-frame video.}
  \vspace{-10pt}
  \label{fig:text_memory}
\end{figure*}

In this section, we outline the pipeline that is employed to generate detailed chain-of-thought traces for the \ourbench training data split.
We begin by generating textual memory traces that detail the interactions performed by the robot during the \textit{Experience Collection} phase (see~\Cref{appendix:experience_collection}). 
For this, we track all the individual actions that are executed by the robot along with necessary information such as time of day, object-recetpacle info, room names that are necessary for successful task completion.
For the co-training experiments for cross-benchmark analysis, we subsample all videos to $48$ frames and accordingly modify the textual memory traces to ensure alignment between the memory trace and subsampled video frames.
A sample original and subsampled textual memory trace is shown in \Cref{fig:text_memory}.

To generate faithful chain-of-thought traces for all samples in the training split, we leverage a zero-shot, text-based LLM to generate a rationale for each task based on the textual memory history and corresponding keyframe solution.
We hypothesize that the zero-shot LLM would be able to generate more consistent traces if it is given access to the solution keyframes and prompted to generate a rationale using the solution.
This is in contrast to using the LLM to predict the solution keyframes based on the textual memory trace which is a more difficult task.
We use the open-source \texttt{DeepSeek-R1-Distill-Qwen-32B}~\citep{guo2025deepseek} for this purpose and leverage the vLLM library~\citep{kwon2023efficient} for accelerated inference. We use the prompt structure as shown in ~\Cref{fig:cot_gen_prompt}. 
After the initial chain-of-thought trace generation, we observed that $\approx8\%$ of the generated traces contained references to the solution keyframe list provided in the prompt. 
To fix these erroneous traces, we use the \texttt{DeepSeek-R1-Distill-Qwen-32B} model to rephrase the generated traces to remove any references to the solution keyframe list following the prompt shown in \Cref{fig:cot_rephrase_prompt}.
After the initial trace generation and rephrasing steps, we are left with $7.8\%$ traces in which the solution keyframes list predicted by the LLM contains a keyframe which does not belong to the actual solution keyframes list.
All these samples are discarded which leads to a total of $72973$ training samples with labeled chain-of-thought traces.

\begin{figure*}[htbp]
\begin{tcolorbox}[
    colback=skyblue!3,          %
    colframe=skyblue!25,        %
    title=\centering{\textcolor{black!100}{\texttt{DeepSeek-R1-Distill-Qwen-32B} CoT Generation Prompt}}, 
    fonttitle=\large,        %
    width=\textwidth,        %
]
\small
You are an intelligent agent assisting a robot in completing specific goals by analyzing interactions recorded during its previous navigation around a house.
The interaction history captures the robot rearranging objects (pick and place).
The robot is given a specific task that needs to be solved by selecting a specific frame index between 0-48.
Each task is solved by selecting the correct frame from the history. \\

Your task is to help the robot generate a detailed reasoning chain of thought to solve the task.
To aid in generating the reasoning trace, you will be given the full list of oracle keyframes that will solve the task. Choosing any frame from this list solves the task. If the oracle list contains only [-1] element, then the task is just unsolvable.
The reasoning\_trace should terminate with a list of predicted solution keyframes. Use the oracle solution to verify your solution\_ keyframes predicted list. \\

You need to solve the following task: \{task\_instruction\}. \\
The oracle solution is: \{oracle\_solution\}. \\
The detailed interactions of the robot are: \{textual\_memory\_trace\} \\

Explanation of oracle\_solution list format: \\
- Single list with multiple sublists. Each sublist can contain multiple frame indices. \\
- If only one sublist, then the task is a single-goal task \\
- If multiple sublists, then the tasks has multiple subgoals and each sublist has corresponds to the solution frames of respective subgoal \\

Please generate the detailed reasoning trace and make sure you refer to the various frame indices and robot actions in the interaction history. \\

Guidelines for reasoning\_trace generation:
- Do not include explicit references to the oracle keyframes. \\
- Your answer should always end with the list of frames that solve the task (use [-1] if unsolvable) based on the reasoning trace. \\
- Ensure the predicted solution\_keyframes follows the format of the oracle\_solution list. \\
- Make sure that each value in the final list of solution\_keyframes you generated exists in the oracle\_solution list that is provided. \\

Please follow the output format:
Output Format (should be in JSON format):
\begin{verbatim}
{
  "reasoning_trace": str.
  "solution_keyframes": str.
}
\end{verbatim}
\end{tcolorbox}
\vspace{-10pt}
\caption{Prompt used for CoT trace generation with \texttt{DeepSeek-R1-Distill-Qwen-32B}.}
\vspace{-10pt}
\label{fig:cot_gen_prompt}
\end{figure*}

\begin{figure*}[htbp]
\begin{tcolorbox}[
    colback=skyblue!3,          %
    colframe=skyblue!25,        %
    title=\centering{\textcolor{black!100}{\texttt{DeepSeek-R1-Distill-Qwen-32B} CoT Trace Rephrasing}}, 
    fonttitle=\large,        %
    width=\textwidth,        %
]
\small
You are given a task instruction and a reasoning trace (chain of thought) that was generated to solve the task. The reasoning trace currently contains explicit reference to ``oracle solution list" or similar oracle-based information that should not be visible to a model trying to solve the task independently. \\

Your job is to rephrase the reasoning trace to remove any explicit references to oracle solutions or oracle keyframes, while preserving the core reasoning logic and thought process. The rephrased reasoning should read as if it was generated by a model thinking through the problem step by step without access to ground truth answers. \\

You should keep references to all other details such as individual frame indices or segments intact as they are necessary for reasoning. \\

Task Instruction: \{task\_instruction\} \\
Original Chain of Thought: \{existing\_chain\_of\_thought\} \\
Please follow the output format:
Output Format (should be in JSON format):
```\begin{verbatim}
{
  "reasoning_trace": str.
}
\end{verbatim}
\end{tcolorbox}
\vspace{-10pt}
\caption{Prompt for CoT Trace Rephrasing with \texttt{DeepSeek-R1-Distill-Qwen-32B}.}
\vspace{-10pt}
\label{fig:cot_rephrase_prompt}
\end{figure*}

\subsection{Sanity Checks}
We investigate how a VLM would perform at short-horizon visual comprehension on images from the same episodes used for long-horizon evaluation in our benchmark. This tests if poor performance of the evaluated models on our long-horizon reasoning tasks is due to a domain or sim2real gap. We assess this by sampling frames around object interactions, asking models to list out detected objects, and check accuracy compared to the ground truth category via string matching -- a conservative metric that misses correct predictions using synonyms or alternative phrasings. This still results in image sequences of length 50-150, but Gemini achieves an accuracy of 67\% which is significantly higher than it's average high level success rate across tasks in \cref{fig:ourbench_main_results}. This shows that despite clear, close-up views, models struggle to reason over them when temporal sequence comprehension is required.

\section{Extended Related Work}
\subsection{An overview of approaches tackling memory in Long VQA and embodied tasks}
\label{related_baselines}
Many recent works equip large language models with memory-like capabilities through task-specific knowledge bases or semantic maps for navigation planning~\citep{2024sif,goat}. MobilityVLA~\citep{xu2024mobility} leverages long-context VLMs to process past frames, enabling goal-frame selection via Gemini, while topological mapping guides navigation without explicit pathfinding. ReMEmbR~\citep{2024arXivremembr} and Embodied RAG~\citep{emb-rag} introduce nonparametric memory trees to store and retrieve past experiences for planning.
For video question answering, \citet{zhang2023simplellovi,2024langrepo} propose a streaming-based approach that accumulates frame-wise captions for later querying. However, this method constrains memory to the expressivity of vision-language or captioning models, potentially omitting critical details. 
We include the hierarchical approach involving long-context VLMs and the approach utilizing textual memory traces as baselines in our evaluations on the benchmark tasks.

\section{Baseline Details} %
\subsection{Gemini Agent}
\label{appendix:gemini_baseline}

\begin{figure*}[htbp]
\vspace{-10pt}
\begin{tcolorbox}[
    colback=skyblue!3,          %
    colframe=skyblue!25,        %
    title=\centering{\textcolor{black!100}{\texttt{gemini-2.0-flash} Task Prompt}}, 
    fonttitle=\large,        %
    width=\textwidth,        %
]
\small
You are an intelligent question answering agent. In the prompt video preceeding the text, you will be shown a long stream of images that have been collected by a robot while navigating around a home and asked to answer a question about the space to assist a user in completing tasks in this home. In the video stream, the robot is picking up objects and rearranging them to different arbitrary locations throughout the house.

The questions asked by user will require being able to look at the full set of images collected by robot to be able to provide the answer.
Your task is to identify the exact point in the video (by timestamp) where the user should move to best accomplish their goal. You can do this by outputting the exact timestep in the video where you are most confident that the object or place that they should move to, was viewed closely.

The images in the video also have the time of day information in top left. You need to use the time of day information for tasks that require to revisit objects or receptacles at a specific time of day.

For some tasks that require revisiting multiple receptacles or objects, you should output multiple frame indices corresponding to the correct order of revisitation (only if specified). For all user goals, identify the minimum number of objects or receptacles (targets) that you need to revisit to complete the user's goal. For each revisitation, you should select one corresponding timestep in the video that you will move to.
You need to ensure that you are not revisiting targets that are not relevant to the user's goal.

Generate timestamps in the \texttt{MM:SS} format where the first two digits represent minutes and the last two digits represent seconds. To solve the tasks effectively, you should try to summarize the video by listing out the objects that were picked and placed, the receptacles from which each object was picked up and the receptacle where it was dropped.

The user's goal is to: \texttt{\{goal\}}. Now look at the full video containing all the images. Give your final response with \texttt{NUM\_TARGETS\_TO\_REVISIT: ||<num\_target>||. TIMESTAMP\_INDEX: ||<timestamp\_1>,\dots,<timestamp\_num\_target>||.} You can include multiple timestamps if the task requires visiting more than one target (one timestamp for each target). Do not add any other text.

\end{tcolorbox}
\vspace{-10pt}
\caption{Prompt for \texttt{gemini-2.0-flash} model}
\vspace{-5pt}
\label{fig:gemini_prompt}
\end{figure*}

We experiment with the \texttt{gemini-2.0-flash} model which can process upto 1 million tokens in the context window. We strictly follow the outlined Google Gemini API developer guidelines~\citep{gemini-api} to structure the prompt and perform API calls over the long videos in the \ourbench tasks. We do not use a ``structured output" format such as json in the output responses as we empirically observed it failed to produce coherent outputs when supplied with long videos in the context. Since the \texttt{gemini-2.0-flash} model expects the input frames to be sampled at 1 FPS, we prompt the model to directly generate specific timestamps in the recommended \texttt{MM:SS} format to localize the frame corresponding to the target entity. We provide the detailed evaluations prompt in Fig. \ref{fig:gemini_prompt}.

\subsection{Qwen, Gemma and GLM Agent}
\label{appendix:qwen_gemma_baseline}

We run the Qwen2.5-VL-7B~\citep{qwen25vl}, Gemma3-12B~\citep{team2025gemma} and GLM-4.1V-Thinking~\citep{hong2025glm} locally for evaluations on the \ourbench tasks. The Qwen model uses a maximum of $768$ frames sampled uniformly across the input sequence while Gemma and GLM does not impose a maximum frame limit. For evaluation on 768 frames, we use two A40 GPUs with 48GB of memory. For experiments with shorter context lengths and higher subsampling (24–384 frames), a single A40 GPU suffices. We provide the full task prompt we use for inference with the three agents in Fig. \ref{fig:qwen_prompt} and utilize structured JSON outputs.

\begin{figure*}[htbp]
\begin{tcolorbox}[
    colback=skyblue!3,          %
    colframe=skyblue!25,        %
    title=\centering{\textcolor{black!100}{\texttt{Qwen/Qwen2.5-VL-7B-Instruct}, \texttt{google/gemma-3-12b-it} and \texttt{zai-org/GLM-4.1V-9B-Thinking} Task Prompt}}, 
    fonttitle=\large,        %
    width=\textwidth,        %
]
\small
You are an expert and intelligent question answering agent. You will be shown a video that was collected by a robot yesterday while navigating around a house and picking and placing objects. 
Your job is to help the robot complete a task today by looking at the video and finding the frame indices that the robot should move to. Make sure your response is all within the JSON. \\

Note:\\
- The robot uses a magic grasp action to pick up an object, where a gripper goes close to the object and the object gets magically picked up.\\

Output Format (should be in JSON format):
\begin{verbatim}
```json
{
  "chain_of_thought": str.
  "frame_indices": list[int].
}
```
\end{verbatim}

Where:\\
- Chain of Thought: A detailed explanation of the thought process in determining the frame indices.\\
- Frame Indices: The frame indices in the video stream showing the object or place that the robot should move to. \\
  - When deciding which frame indices to choose, make sure you choose the frame indices that are closest to the object/place. \\
  - If task requires the agent to go to multiple places, output one frame index per object/place. (Do not use ellipsis)\\

The robot's goal is: \texttt{\{goal\}}

\end{tcolorbox}
\vspace{-10pt}
\caption{Prompt for \texttt{Qwen2.5-VL} and \texttt{Gemma3} model}
\vspace{-5pt}
\label{fig:qwen_prompt}
\end{figure*}

\begin{figure*}[htbp]
\begin{tcolorbox}[
    colback=skyblue!3,          %
    colframe=skyblue!25,        %
    title=\centering{\textcolor{black!100}{\texttt{gpt-4o-2024-07-18} Task Prompt}}, 
    fonttitle=\large,        %
    width=\textwidth,        %
]
\small
** OBJECTIVE **\\
You are an expert and intelligent question answering agent. You will be shown a video that was collected by a robot yesterday while navigating around a house and picking and placing objects. \\
Your job is to help the robot complete a task today by looking at the video and finding the frame indices that the robot should move to, to complete the task.\\

** OUTPUT FORMAT **
\begin{verbatim}
```json
  "chain_of_thought": str.
  "frame_indices": list[int].
```
\end{verbatim}

Where:\\
- Chain of Thought: A detailed explanation of the thought process for determining the frame indices.\\
- Frame Indices: The frame indices in the video stream showing the object or place that the robot should move to.\\
  - When deciding which frame indices to choose, make sure you choose the frame indices that are closest to the object/place.\\
  - If task requires the agent to go to multiple places, output one frame index per object/place.\\

** NOTES **\\
- The robot uses a magic grasp action to pick up an object, where a gripper goes close to the object and the object gets magically picked up.\\

** TASK **\\
The robot's goal is: \texttt{\{goal\}}
\end{tcolorbox}
\vspace{-10pt}
\caption{Prompt for \texttt{gpt-4o} model}
\vspace{-10pt}
\label{fig:gpt_prompt}
\end{figure*}

\subsection{GPT Agent}
\label{appendix:gpt_baseline}

We use the official OpenAI GPT-4o~\citep{hurst2024gpt} API to run evaluations on the \ourbench tasks with video frames subsampled at $96$ frames to optimize API costs. For running complete evaluation on the entire validation task suite, we incurred a total cost of $\approx400$ USD. We provide the full system prompt used for GPT-4o evaluations in Fig. \ref{fig:gpt_prompt}.

\subsection{Text Agent}
\label{appendix:text_agent}

We implement the the text-based VLM agent using the Qwen2.5-VL~\citep{qwen25vl} model. For every \texttt{chunk\_size} frames in the video, we generate a description using the VLM. We then attach the frame indices and \textit{time of day} information to the generated summary for the frame window in consideration. We use the prompt shown in Fig. \ref{fig:text_vlm_prompt} to generate the structured JSON outputs for each chunk in the video. We then concatenate all the generated JSON outputs and pass it to the Qwen model to perform text-only reasoning over the entire text-based history of the video. The text-based reasoning prompt is provided in Fig. \ref{fig:final_llm_prompt}. We note that using the same Qwen model for generating the individual text summaries and performing the final text-based reasoning helps in optimizing inference load as we maintain only a single instance of the VLM through the entire evaluation.  

\begin{figure*}[htbp]
\begin{tcolorbox}[
    colback=skyblue!3,          %
    colframe=skyblue!25,        %
    title=\centering{\textcolor{black!100}{\texttt{Text Agent(VLM)} Task Prompt}}, 
    fonttitle=\large,        %
    width=\textwidth,        %
]
\small
You are an expert and intelligent question answering agent. You will be shown a stream of \texttt{\{chunk\_size\}} images taken by a robot while navigating around a simulated house and picking and placing objects. Your job is to describe the image based on the output format. Make sure your response is all within the JSON format. \\

Output Format (should be in JSON format):
\begin{verbatim}
```json
{
  "room_name": str.
  "picking_placing_or_navigating": str("picking", "placing" or "nav").
  "object_being_manipulated": str.
  "receptacle_being_manipulated": str.
  "other_objects_in_scene": list[str].
}
```
\end{verbatim}

Where: \\
- Room name: The name of the room the robot is in. \\
- Picking, placing, or navigating: Whether the robot is picking or placing an object, or navigating around the house. \\
- Object being manipulated: The object that the robot is picking or placing (if relevant). \\
- Receptacle being manipulated: The receptacle that the robot is picking up from or placing an object into (if relevant). \\
- Other objects in scene: Objects present in the scene besides the main object and the receptacle. \\

Note: \\
- The robot uses a magic grasp action to pick up an object, where a gripper goes close to the object and the object gets magically picked up.
\end{tcolorbox}
\vspace{-10pt}
\caption{Prompt to generate text summaries for \texttt{chunk\_size} frames with \texttt{Qwen2.5-VL}}
\vspace{-10pt}
\label{fig:text_vlm_prompt}
\end{figure*}

\begin{figure*}[h]
\begin{tcolorbox}[
    colback=skyblue!3,          %
    colframe=skyblue!25,        %
    title=\centering{\textcolor{black!100}{\texttt{Text Agent(LLM)} Task Prompt}}, 
    fonttitle=\large,        %
    width=\textwidth,        %
]
\small
You are an expert and intelligent question answering agent. You will be provided a list of textual descriptions of the environment created by an agent while navigating around a house and picking and placing objects. The descriptions correspond to logs made after collecting and viewing a small chunk of frames, and contain information about which frame numbers they correspond to. You will also be provided a goal that now needs to be accomplished, for which you will need to use the history to decide where to go. Your job is to identify the desired frame index to navigate to, based on the provided task. Make sure your response is all within the JSON. \\

Agent's History information: \\
- Room name: The name of the room the agent was in. \\
- Picking, placing, or navigating: Whether the robot was picking or placing an object, or navigating around the house. \\
- Object being manipulated: The object that the robot was picking or placing (if relevant). \\
- Receptacle being manipulated: The receptacle that the robot was picking up from or placing an object into (if relevant). \\
- Other objects in scene: Objects present in the scene besides the main object and the receptacle. \\

Output Format (should be in JSON format):
\begin{verbatim}
```json
{
  "chain_of_thought": str.
  "frame_indices": list[int].
}
```
\end{verbatim}

Where: \\
- Chain of Thought: A detailed explanation of your thought process in determining the frame indices. \\
- Frame Indices: A list of one or more frame indices relevant to accomplishing the goal. \\
  - If its a single object/place task, output only one frame index. \\
  - If task requires the agent to go to multiple places, output one frame index per object/place. (Do not use ellipsis) \\

Agent's history: \{history\} \\

The goal is \texttt{\{goal\}}
\end{tcolorbox}
\vspace{-10pt}
\caption{Prompt to perform reasoning over the combined textual summaries with \texttt{Qwen2.5-VL}}
\vspace{-10pt}
\label{fig:final_llm_prompt}
\end{figure*}

\subsection{Supervised Finetuning Baseline}
\label{appendix:sft_baseline}

We leverage the \ourbench training split which consists of $82174$ unique tasks across the categories mentioned in \Cref{tab:task_categories}. 
Each video-instruction pair is coupled with a list of ground-truth frame indices that solve the task. For multi-goal tasks, each subgoal has a corresponding ground-truth frame indices sublist. 
Since the \ourbench video sequences can be extremely long, we uniformly subsample all videos to $96$ frames. The ground truth frame indices are also subsampled to match the shorter video lengths to ensure the model chooses from within the subsampled frames.

\textbf{Training Details.} We conduct full-finetuning of the \texttt{Qwen/Qwen2.5-VL-3B} model using the open-source \texttt{huggingface/trl} library.
For hyperparameter optimization, we utilized a smaller representative split of the training dataset with $36274$ samples and conducted a grid search over the learning rate, num\_epochs and weight\_decay parameters.
The best checkpoints from all runs were selected by evaluating performance on all video-instruction pairs in the full validation split.
This ``offline" evaluation is conducted without instantiating the simulator by directly comparing the model's predicted keyframe indices with the ground-truth keyframe indices using exact string matching (thus, we dont compute metrics proposed in \Cref{sec:eval_metrics}).
We use the cosine learning rate decay scheduler for all experiments.
For the final training on the larger dataset, we found that a learning rate of $5 \times 10^{-6}$ over $5$ epochs with $0$ weight decay worked best.
For all experiments, we train in bfloat16 format using 8 A40 GPUs on a single node with a batch size of $1$ and gradient\_accumulation of $4$ - providing an effective batch\_size of $32$ samples.
To further optimize training, we use gradient\_checkpointing and highly optimized flash\_attention implementation. In this setup, the full training run takes $\approx 120$ hrs to complete.

\textbf{Goal Sampling for Online Simulator Eval.} The best trained checkpoint is used in online simulator evaluations on the \ourbench task suite.
Since the model is trained to predict the complete list of ground-truth frame indices, we select a single representative index from each generated subgoal solution list to serve as the predicted frame index. 
Empirically, we found that choosing either the first or a random index (of a sublist corresponding to a subgoal) yields comparable performance, while using the last index results in slightly worse performance. Therefore, we use the first index of each list as a simple and effective heuristic.

\section{Low-level Policy Details}

\subsection{Image-Goal Navigation Agent}
\label{sec:imagenav_details}

\textbf{Policy Training.} 
We train the image-goal navigation policy using the architecture and training recipe described in ~\citet{yadav2023ovrlv2}. Specifically, we employ an end-to-end reinforcement learning policy trained using DDPPO, which predicts discrete navigation actions conditioned on visual RGB inputs and a goal image. The agent utilizes a ViT based visual encoder along with a 2-layer LSTM backbone. Similar to OVRL-v2~\citep{yadav2023ovrlv2}, the output patch representations from the ViT are reshaped into a 3D grid and downsampled to a lower dimension using a convolutional layer called the compression layer. However, we slightly deviate from OVRL-v2 by first concatenating the patch representations from the current and goal images before passing them through the compression layer.

The policy training occurs in the training scene split of the HSSD dataset~\citep{hssd}, consisting of $1166$ episodes distributed across $111$ scenes (a subset of the original $125$ training scenes). In the Habitat simulator, the agent is modeled as a Hello Robot Stretch~\citep{kemp2022design} with a height of $1.41$ m and a cylindrical base radius of $0.25$ m. The agent’s RGB sensor is positioned at a height of $1.31$ m, with a resolution of $160 \times 120$ pixels and a horizontal field of view of $43^\circ$. During training, each episode has a maximum step budget of $1000$ steps, with success defined by invoking the \texttt{STOP} action within $1.0$ m of the goal image position.

As our visual encoder, we use the VC-1-Base~\citep{majumdar2023we} model which was previously finetuned with ImageNav on a smaller training scene split from the HSSD dataset. Empirically, we found that freezing the visual encoder after this targeted finetuning significantly accelerates the overall training performance. During the VC-1 finetuning, we use a smaller learning rate of $1.5 \times 10^{-6}$ for the encoder.

We train the agents for a total of $500$M timesteps on 32 A40 GPUs running 32 parallel environments each (1024 envs in total). Following the approach from ~\citep{majumdar2023we}, we collect $64$ steps of experience and subsequently perform 2 PPO epochs with a mini-batch size of $2$. The reward function employed is the improved formulation proposed by~\citet{yadav2023ovrlv2}, using parameters: success weighting $c_s = 5.0$, angle success weighting $c_a = 5.0$, goal radius $r_g = 1.0$, angle threshold $\theta_g = 25^\circ$, and slack penalty $\gamma = -0.002$. We optimize using AdamW with a learning rate of $2.5 \times 10^{-4}$ and weight decay $10^{-6}$.

We select the best-performing policy checkpoint based on a validation success rate obtained by evaluating on $1000$ image-goal navigation episodes from the \ourbench validation set, achieving a success rate of 78\% and SPL of 48.21\%.

\subsection{Explicit Mapping Agent}
\label{sec:mapper_details}
\label{sec:hierarchical_agent}
\vspace{-5pt}
\begin{figure*}[!htb]
  \centering
  \includegraphics[width=\linewidth]{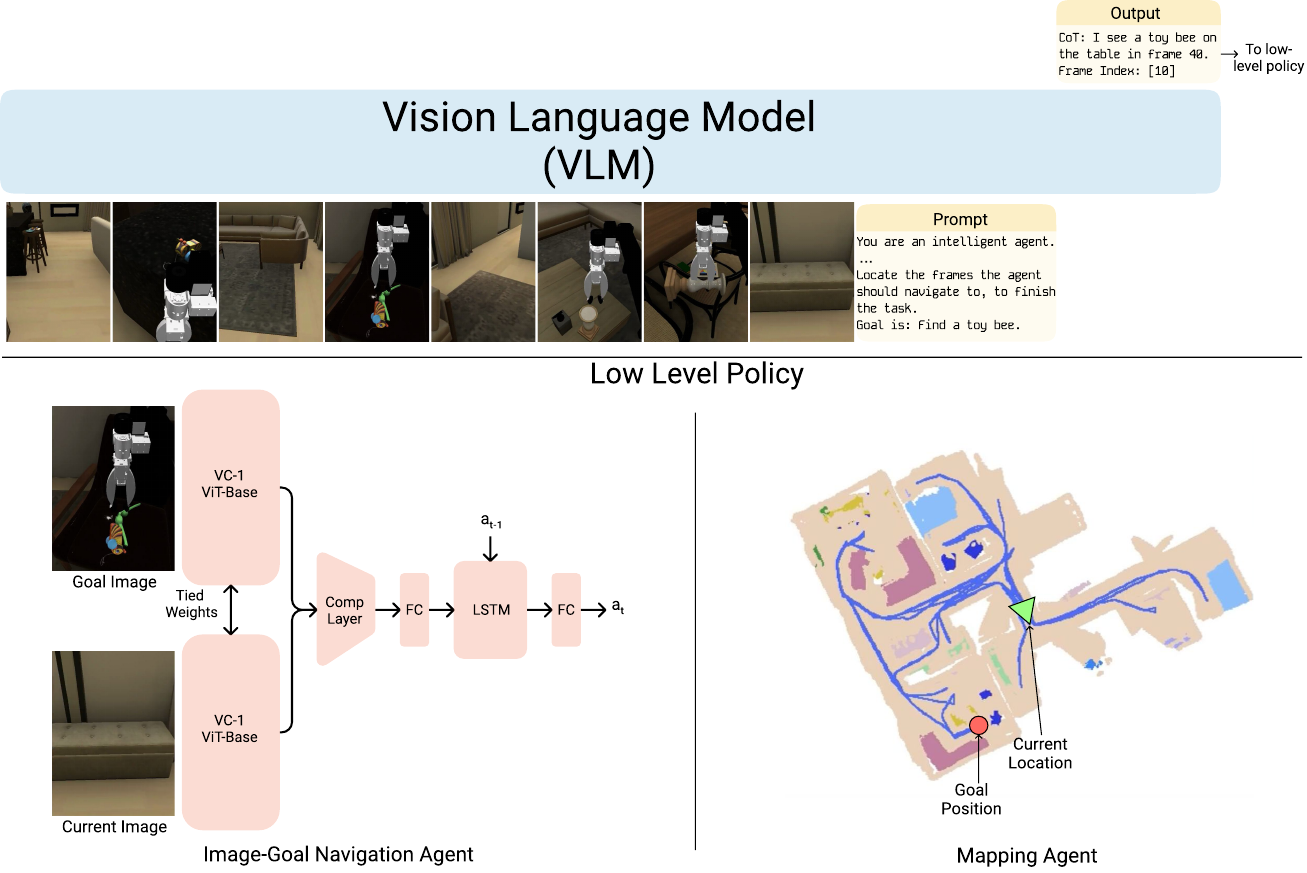}
  \vspace{-10pt}
  \caption{Hierarchical agent architecture when using a VLM for goal frame selection and low level policies for action generation.}
  \vspace{-10pt}
  \label{fig:ourbench_agent}
\end{figure*}

We augment the \texttt{OracleAgent} (see \Cref{appendix:experience_collection}) with a top-down occupancy map as it navigates the environment to generate the video sequence. In addition, we also store the 2D poses corresponding to each frame that is collected in the video sequence. The constructed map is employed by a path planning module to generate a deterministic navigation sequence to the 2D map coordinates that correspond to the image-goal location selected by the high-level goal selection agent (see \Cref{sec:hierarchical_baseline}).

\textbf{Map Construction.} We leverage the onboard depth sensor and project the depth image at each timestep into an egocentric pointcloud using camera intrinsics. The pointcloud is then binned along the z-axis to compute occupancy values on a local 2D grid map centered around the current position of the agent. The pose sensor provides the instantaneous pose of the agent which is employed to register the computed local map in the global map of the environment. In this way, we continuously update the global map as the \texttt{OracleAgent} navigates the environment and visits multiple pick-place targets. We freeze the map updates during the pick-place subroutines as no additional occupancy information is encountered during their execution.

\textbf{Planner.} When the high-level goal agent selects a particular goal frame, we lookup the corresponding target 2D pose in the global map. Similar to the local navigation policy employed in~\citet{goat}, we use the Fast Marching Method~\citep{fmm} to generate the shortest path to the goal pose. We use one of four discrete navigation actions (see \Cref{sec:implementation_details}) to reach each waypoint on the generated shortest path.

\section{Cross-Benchmark Experiments}
\label{appendix:cross_benchmark_exps}

In this section, we provide details about the training and evaluation setup for results on the \texttt{VSI-Bench} \cite{yang2024thinking} dataset. 
For evaluation, we follow the exact same evaluation protocol outlined in the original paper.
In ~\cref{tab:vsi_bench_results}, we report results across the multiple-choice ($2490$ in number) and numerical regression ($2640$ in number) QA categories, each of which has four sub-categories.
For all experiments, we use the \texttt{Qwen2.5-VL-3B} as the base model.
Each SFT/RL-finetuning experiment uses a single node of $8$ A40 GPUs with $48$ GB VRAM each.

\textbf{Supervised Finetuning. } We conduct supervised finetuning of model with maximum of $48$ frames in the input context window.
We also use $48$ frames during model inference as that works best between $32$, $48$ and $64$ frame inference for all baselines.
For SFT experiments with the \texttt{Video-R1-CoT-165k}-\ourbench mixture, we experimented with various mixing ratios by changing the amount of \ourbench data ($12.5\%, 25\%, 37.5\%, 45\%$) and found $25\%$ ($40000$ samples) to work best.
All SFT models are trained for one epoch following the hyperparameters from \cite{feng2025video}. 
We only evaluate the final checkpoint after one epoch for all baselines.
Specifically, we use a learning rate of $1 \times 10^{-6}$, gradient\_accumulation of 2, max\_grad\_norm of 5 and train with bfloat16 precision on 8 GPUs
We leverage gradient\_checkpointing, flash\_attention and deepspeed\_zero\_2 to optimize training.

\textbf{RL Finetuning. }
For RL finetuning with GRPO, we reduce the context window to $32$ frames to avoid out-of-memory errors encountered with $48$ frames. 
Emperically, we observe that inferencing with $64$ frames works better than $32$ frame inference for all baselines.
We mix the full \ourbench training dataset ($\sim78$k samples) with the \texttt{Video-R1-260k} dataset from \citet{feng2025video}. 
Because each training step is computationally expensive, we train for only $1000$ gradient steps using a group size of $8$, batch size of $1$ per GPU, and gradient accumulation of $2$. 
As a result, the model is exposed to only $\sim2000$ unique samples during training. 
We set the weight decay to $0.01$, maximum gradient norm to $5$, learning rate to $1\times10^{-6}$, and maximum completion length to $768$. 
We evaluate checkpoints at a $100$ step frequency and select the checkpoint with highest average MCQ+Numerical accuracy for each baseline.

For \ourbench samples, we design a custom reward function that rewards the precision of the predicted goal frames. Let $\mathcal{R}=(R_1,\ldots,R_n)$, $\mathcal{H}=(H_1,\ldots,H_m)$, where $R_i \subset \mathbb{Z}$ (reference frame sets) and $H_i$ are finite sequences of predicted frames.

Define the score $S(\mathcal{H},\mathcal{R})$ as
\[
S(\mathcal{H},\mathcal{R}) =
\begin{cases}
0, & \text{if } n \neq m,\\[6pt]
\dfrac{\displaystyle \sum_{i=1}^{n} \sum_{h \in H_i} \mathbf{1}\!\left[h \in R_i\right]}
{\displaystyle \sum_{i=1}^{n} |H_i|}, & \text{otherwise (assuming } \sum_{i=1}^{n}|H_i|>0\text{).}
\end{cases}
\]

\section{Qualitative Examples}
\label{appendix:qual_examples}

\begin{figure}[htb]  %
    \centering
    \includegraphics[width=\textwidth]{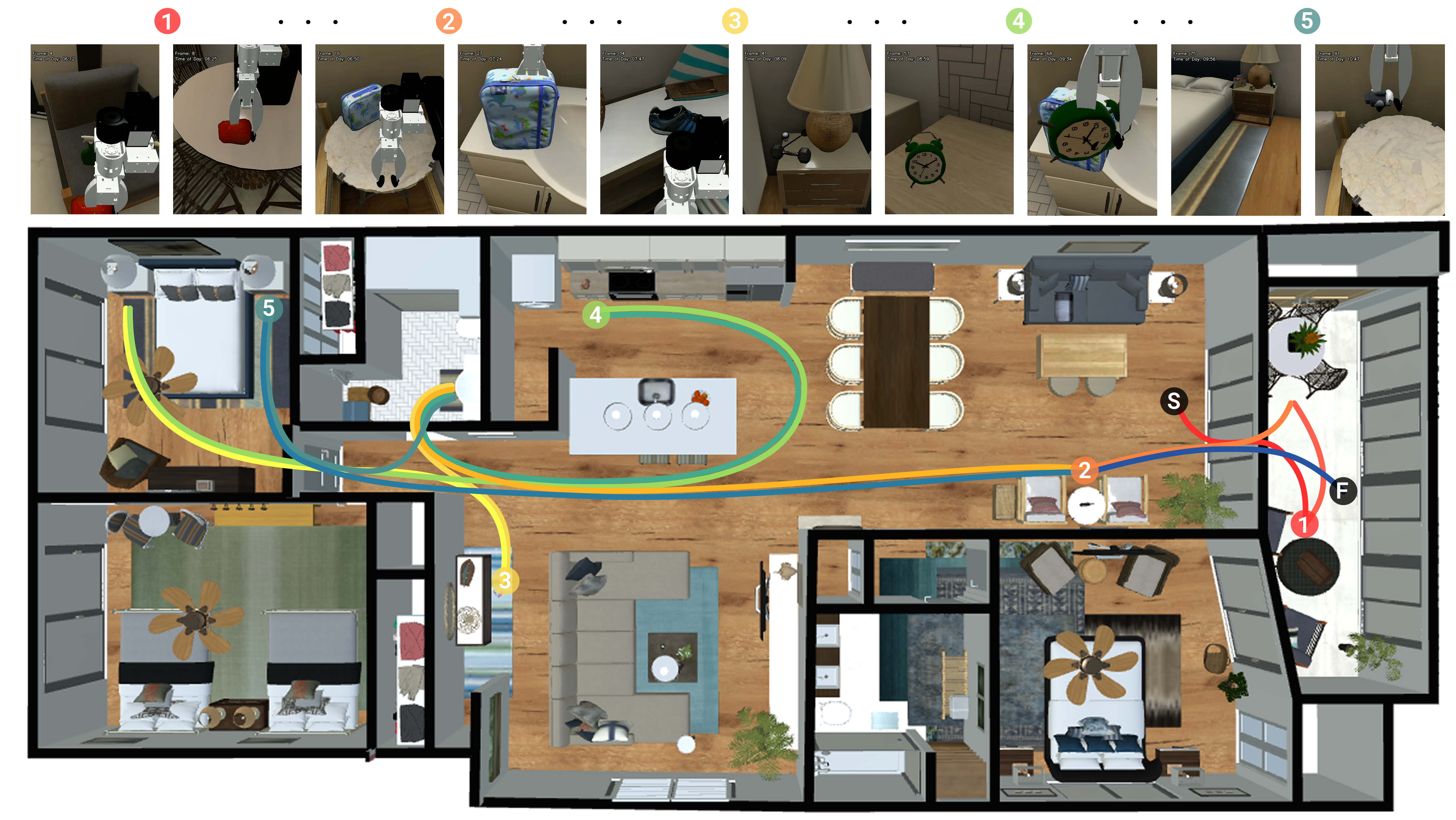}  %
    \vspace{-20pt}
    \caption{An example experience collection trajectory executed by \texttt{OracleAgent}.}
    \vspace{-10pt}
    \label{fig:qualitative_traj}
\end{figure}

\begin{figure}[htbp]  %
    \centering
    \includegraphics[width=\textwidth, height=0.99\textheight]{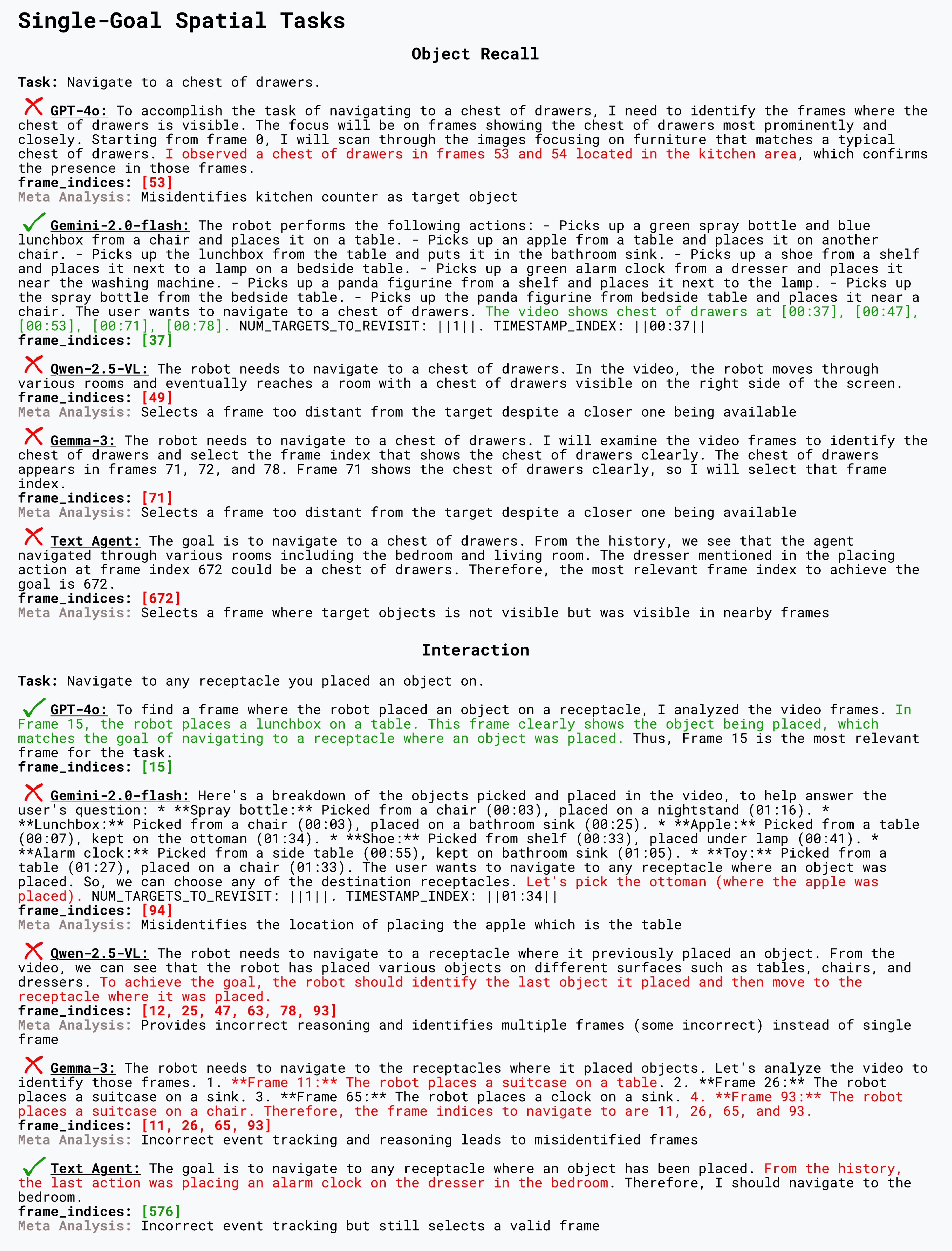}
    \caption{\ourbench Spatial Tasks for the sample episode in \Cref{fig:qualitative_traj}. Each VLM model uses $96$ subsampled frames. \textit{Text Agent} processes full video as defined in \cref{appendix:text_agent}.}
    \label{fig:spatial_ex_1}
\end{figure}

\begin{figure}[htbp]  %
    \centering
    \includegraphics[width=\textwidth, height=0.99\textheight]{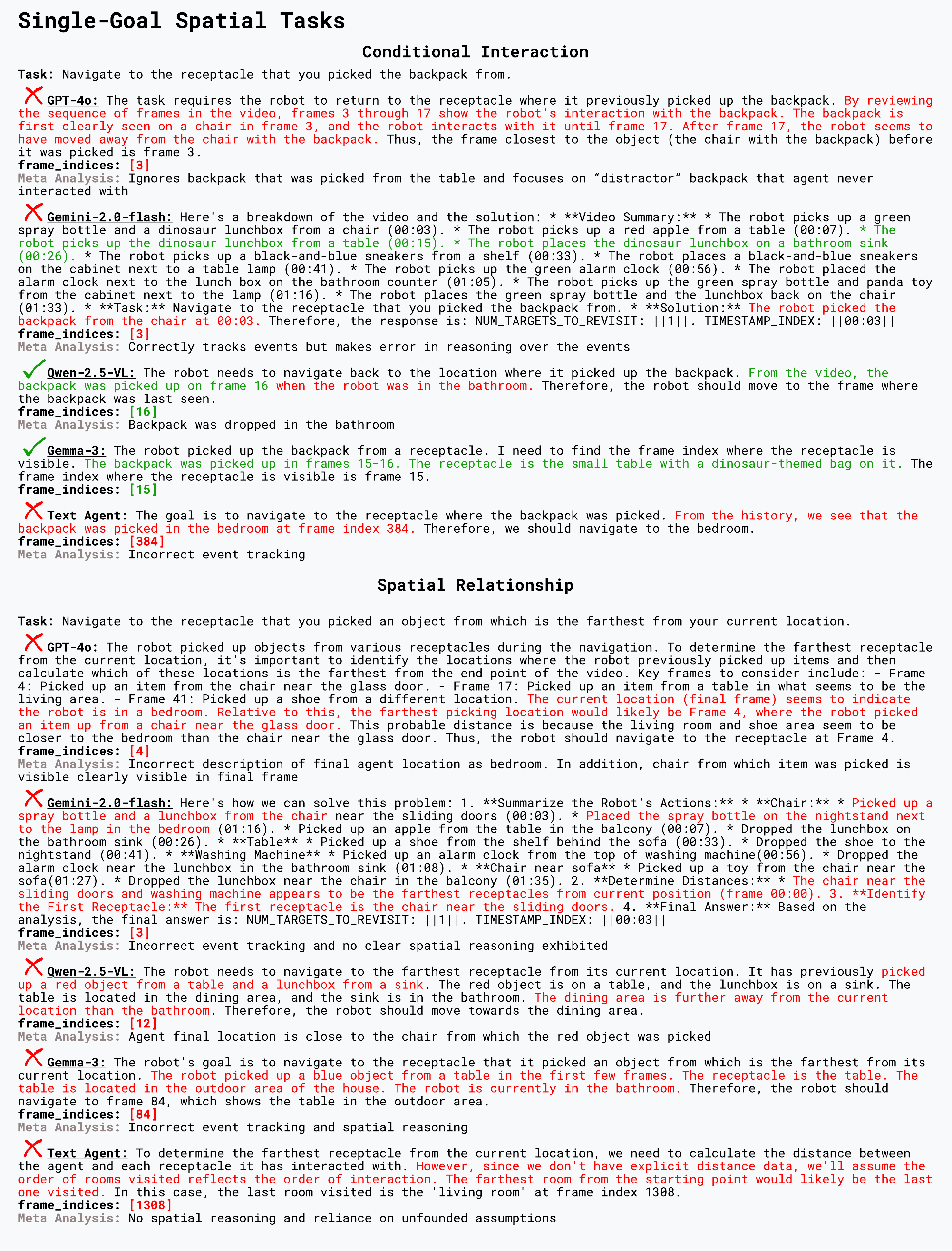}
    \caption{\ourbench Spatial Tasks for the sample episode in \Cref{fig:qualitative_traj}. Each VLM model uses $96$ subsampled frames. \textit{Text Agent} processes full video as defined in \cref{appendix:text_agent}.}
    \label{fig:spatial_ex_2}
\end{figure}

\begin{figure}[htbp]  %
    \centering
    \includegraphics[width=\textwidth, height=0.99\textheight]{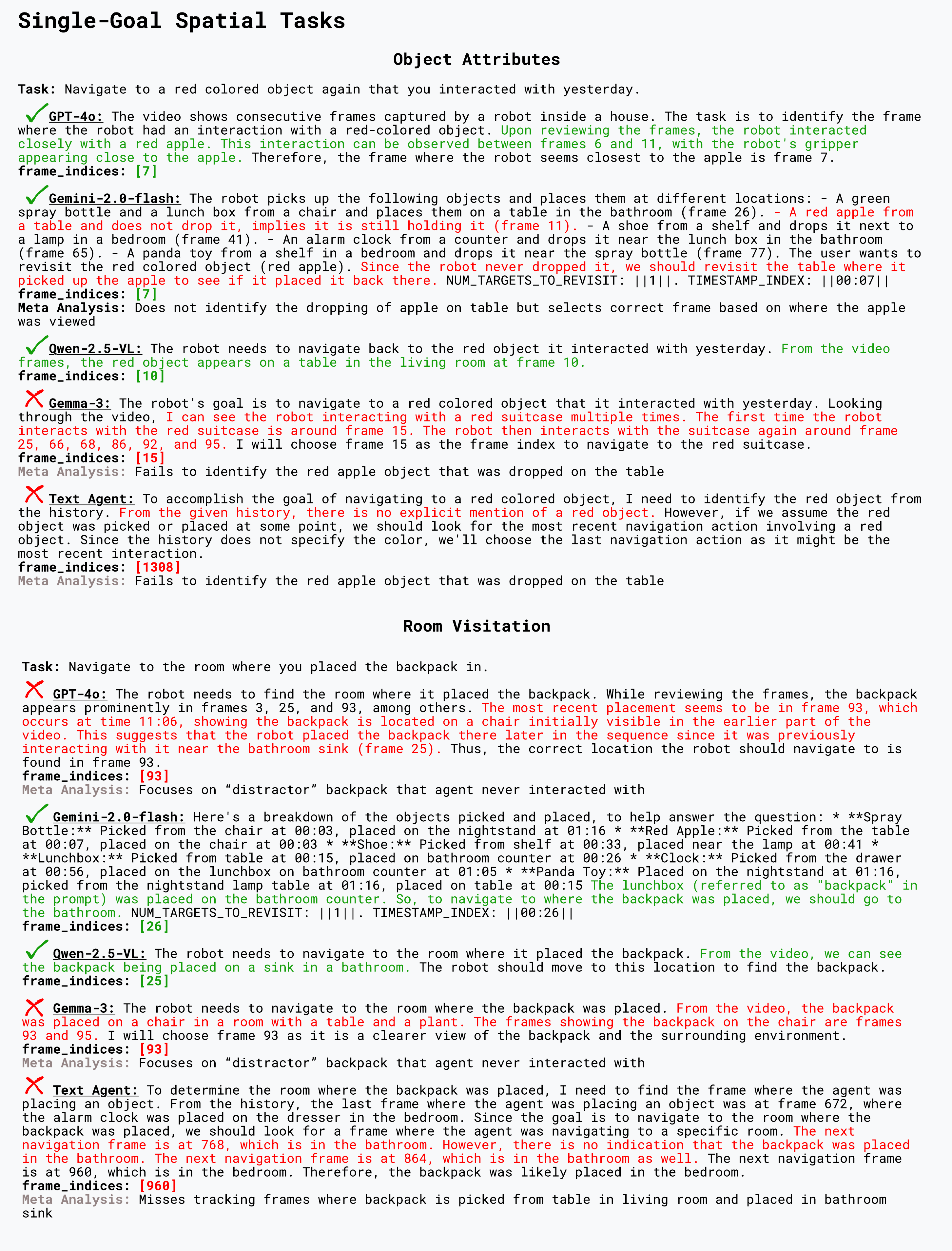}
    \caption{\ourbench Spatial Tasks for the sample episode in \Cref{fig:qualitative_traj}. Each VLM model uses $96$ subsampled frames. \textit{Text Agent} processes full video as defined in \cref{appendix:text_agent}.}
    \label{fig:spatial_ex_3}
\end{figure}

\begin{figure}[htbp]  %
    \centering
    \includegraphics[width=\textwidth, height=0.99\textheight]{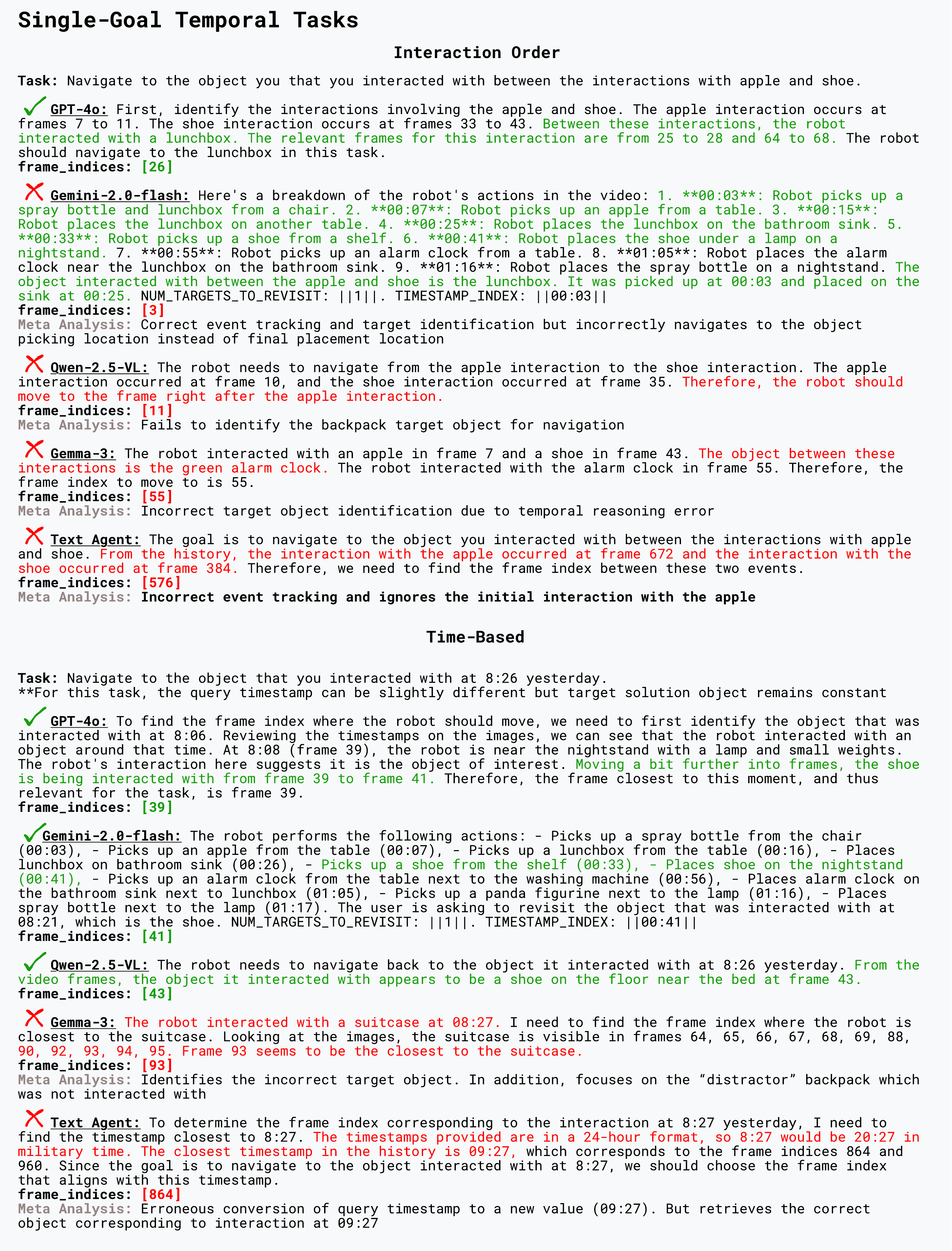}
    \caption{\ourbench Temporal Tasks for the sample episode in \Cref{fig:qualitative_traj}. Each VLM model uses $96$ subsampled frames. \textit{Text Agent} processes full video as defined in \cref{appendix:text_agent}.}
    \label{fig:temporal_ex_1}
\end{figure}

\begin{figure}  %
    \centering
    \includegraphics[width=\textwidth]{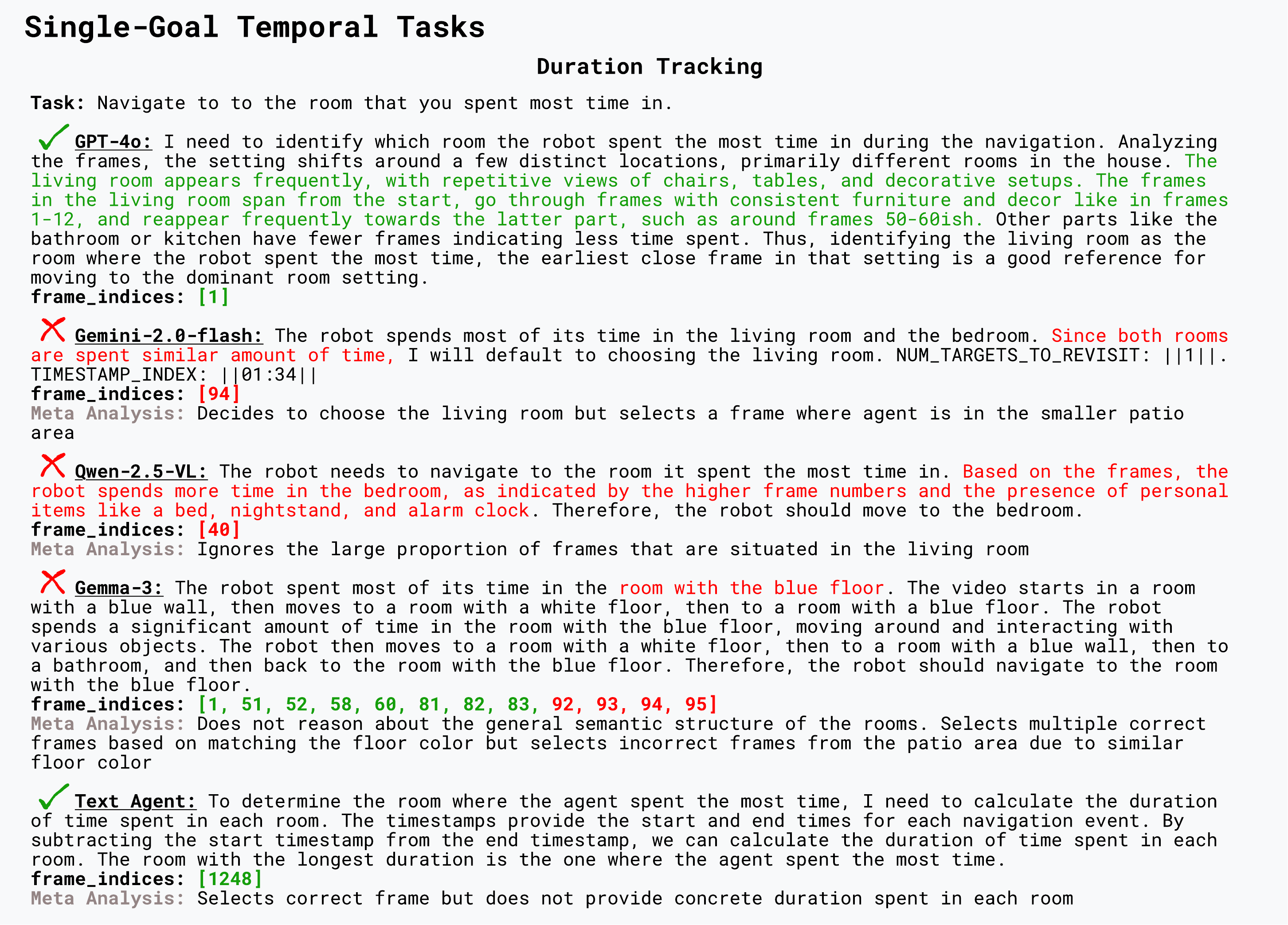}
    \caption{\ourbench Temporal Tasks for the sample episode in \Cref{fig:qualitative_traj}. Each VLM model uses $96$ subsampled frames. \textit{Text Agent} processes full video as defined in \cref{appendix:text_agent}.}
    \label{fig:temporal_ex_2}
\end{figure}

\begin{figure}[htbp]  %
    \centering
    \includegraphics[width=\textwidth, height=0.99\textheight]{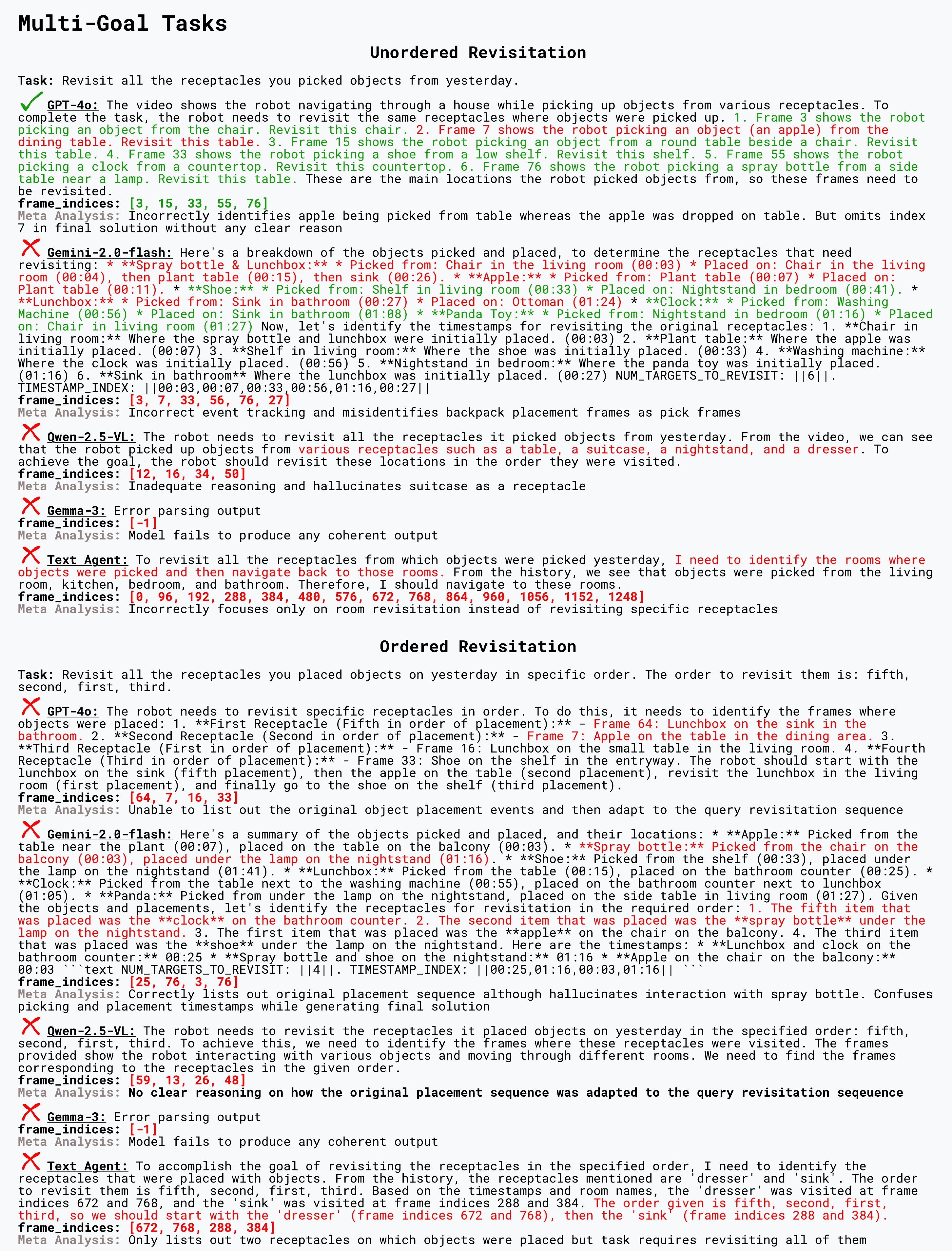}
    \caption{\ourbench Multi-Goal Tasks for the sample episode in \Cref{fig:qualitative_traj}. Each VLM model uses $96$ subsampled frames. \textit{Text Agent} processes full video as defined in \cref{appendix:text_agent}.}
    \label{fig:multigoal_ex}
\end{figure}

In this section, we include representative qualitative examples from the \ourbench task suite. In \Cref{fig:qualitative_traj}, we provide a birds-eye view depiction of the trajectory executed during experience collection (see \Cref{appendix:experience_collection}) in a validation episode involving 5 pick-place interaction routines. For this episode, the collected video consists of $1308$ frames in total. We provide detailed responses (and meta-analysis) from various baselines across the various task categories in \Cref{fig:spatial_ex_1,fig:spatial_ex_2,fig:spatial_ex_3,fig:temporal_ex_1,fig:temporal_ex_2,fig:multigoal_ex}. In the following, we briefly discuss the different failure modes encountered in the generated responses across the 3 high-level task categories defined in \Cref{tab:task_categories}.

\textbf{Single-Goal Spatial Tasks.} We observe that for the \textit{Object Recall} task (\cref{fig:spatial_ex_1}), the models are able to identify a frame with the target \textit{chest of drawers} but do not select the closest frames to ensure success. In the \textit{Interaction} based tasks (\cref{fig:spatial_ex_1,fig:spatial_ex_2}), we observe models fail to accurately list the sequence of interaction events which can be partially attributed to supplying only $96$ subsampled frames from the original video. In one case, \texttt{GPT-4o} instead focuses on a ``distractor" target object (\cref{fig:spatial_ex_2}). All models perform poorly on the \textit{Spatial Relationship} tasks (\cref{fig:spatial_ex_2}) as they are unable to coherently reason about the spatial layout of the house in the provided video. 

\textbf{Single-Goal Temporal Tasks.} For the \textit{Interaction Order} task, we observe that the closed-source models are able to accurately identify the target object but the open-source variants fail to do so (\cref{fig:temporal_ex_1}). Models are generally able to select the correct frame in the \textit{Time-Based} task as the model only needs to retrieve a single frame corresponding to the correct overlaid timestamp. For the \textit{Duration Tracking} task, only \texttt{GPT-4o} generates the correct reasoning and frame prediction. Intersetingly, \texttt{Gemma-3} tries to track the duration spent in each room by matching the floor color observed in the images but eventually fails at the task (\cref{fig:temporal_ex_2}).

\textbf{Multi-Goal Tasks.} All models fail to perform successfully on tasks under this category. Only \texttt{GPT-4o} gets a correct response for all subgoals. Surprisingly, the reasoning trace has a hallucination but this mistake is omitted in the final frame sequence prediction (\cref{fig:multigoal_ex}). The models struggle to coherently list the event sequence and also hallucinate interactions. Furthermore, \texttt{Gemini-2.0-flash} also struggles to adapt the sequence of events it lists into the query revisitation sequence specified in the task (\cref{fig:multigoal_ex}).

\begin{table}[!htbp]
\vspace{-10pt}
\centering
\scriptsize
\begingroup
\setlength{\tabcolsep}{3pt} %
\renewcommand{\arraystretch}{1.12} %
\resizebox{\linewidth}{!}{%
\begin{tabular}{lrrrrrrrrrrrr}
\toprule
Method &
\begin{tabular}[c]{@{}c@{}}All\\Tasks\end{tabular} &
\begin{tabular}[c]{@{}c@{}}Object\\Recall\end{tabular} &
\begin{tabular}[c]{@{}c@{}}Inter-\\action\end{tabular} &
\begin{tabular}[c]{@{}c@{}}Conditional\\Interaction\end{tabular} &
\begin{tabular}[c]{@{}c@{}}Object\\Attributes\end{tabular} &
\begin{tabular}[c]{@{}c@{}}Spatial\\Relation\end{tabular} &
\begin{tabular}[c]{@{}c@{}}Room\\Visitation\end{tabular} &
\begin{tabular}[c]{@{}c@{}}Interaction\\Order\end{tabular} &
\begin{tabular}[c]{@{}c@{}}Time-\\Based\end{tabular} &
\begin{tabular}[c]{@{}c@{}}Duration\\Tracking\end{tabular} &
\begin{tabular}[c]{@{}c@{}}Unordered\\Revisitation\end{tabular} &
\begin{tabular}[c]{@{}c@{}}Ordered\\Revisitation\end{tabular} \\
\midrule
\rowcolor{gray!15}
Oracle           & 95.92 & 100.00 & 98.67 & 92.03 & 99.80 & 85.42 & 80.00 & 99.86 & 99.50 & 100.00 & 99.83 & 99.00 \\
Text Agent       &  9.45 &   9.50 & 20.00 &  7.99 &  6.84 &  4.17 & 19.36 &  7.37 & 24.50 &  15.46 &  0.83 &  0.67 \\
Gemma3           & 13.04 &  19.50 &  9.00 & 13.22 & 17.82 &  5.83 & 30.61 & 12.31 & 28.50 &  21.13 &  0.33 &  2.00 \\
Qwen2.5-VL       & 15.14 &  15.50 & 35.00 & 15.22 & 17.62 &  4.58 & 31.83 &  9.83 & 40.00 &  12.76 &  1.83 &  1.33 \\
GLM-4-1v         & 23.49 &  29.00 & 49.17 & 24.59 & 30.35 &  7.08 & 36.34 & 21.05 & 50.50 &  16.11 &  3.33 &  3.00 \\
Gemini-2.0-Flash & 25.73 &  47.00 & 33.83 & 30.16 & 35.27 & 10.42 & 41.01 & 23.23 & 43.50 &  21.48 &  8.50 &  9.33 \\
GPT-4o           & 27.33 &  32.00 & 40.00 & 28.19 & 25.20 & 14.58 & 44.68 & 28.56 & 63.00 &  24.83 &  6.33 &  7.00 \\
Qwen SFT         & 52.44 &  64.50 & 70.63 & 46.14 & 42.65 & 24.58 & 53.47 & 68.22 & 88.50 &  53.53 & 29.70 & 20.00 \\
\bottomrule
\end{tabular}%
}
\endgroup
\vspace{-10pt}
\caption{High-level goal success rate (\%) across task categories. The \textit{Oracle} row is shaded in {\color{gray}gray}.}
\label{tab:goal_success}
\vspace{-10pt}
\end{table}

\begin{figure}[htb]  %
    \centering
    \includegraphics[width=\textwidth]{figures/pick_place.pdf}  %
    \vspace{-10pt}
    \caption{Example ``magic" pick-place sequences executed by the \texttt{OracleAgent} for sample episode in Fig. \ref{fig:qualitative_traj}.}
    \vspace{-10pt}
    \label{fig:oracle_pick_place}
\end{figure}

\clearpage

\small  %
\setlength{\tabcolsep}{2pt}  %
\renewcommand{\arraystretch}{1.1}  %

\begin{longtable}{l p{0.8\linewidth}}
    \caption{Task Categories with all associated prompts}
    \label{tab:all_tasks} \\
    \toprule
    \rowcolor{lightgray!20} 
    \textbf{Memory Type} & \textbf{Example Instructions} \\
    \midrule
    \endfirsthead
    
    \toprule
    \rowcolor{lightgray!20} 
    \textbf{Memory Type} & \textbf{Example Instructions} \\
    \midrule
    \endhead

    \bottomrule
    \multicolumn{2}{r}{\textit{Continued on the next page...}} \\
    \endfoot

    \bottomrule
    \endlastfoot

    \graycell
    & \graycell Navigate to a \{target\_object\_name\}. \\
    \graycell \multirow{-2}{*}{\textbf{Object Recall}} 
    & Navigate to a \{target\_receptacle\_name\}. \\

    & \graycell Navigate to any receptacle you interacted with. \\
    & Navigate to any receptacle you did not interact with. \\
    & \graycell Navigate to any object that you interacted with yesterday. \\
    & Navigate to any object that you did not interact with yesterday. \\
    & \graycell Navigate to any receptacle you picked an object from. \\
    \multirow{-6}{*}{ \textbf{Interaction}} 
    & Navigate to any receptacle you placed an object from. \\    

    \graycell & \graycell Navigate to the \{target\_object\_name\} that you interacted with yesterday. \\
    \graycell & Navigate to a \{target\_object\_name\} that you did not interact with yesterday. \\
    \graycell & \graycell Navigate to a \{target\_receptacle\_name\} you did not interact with yesterday. \\
    \graycell & Navigate to a \{target\_receptacle\_name\} you picked an object from. \\
    \graycell & \graycell Navigate to a \{target\_receptacle\_name\} you placed an object on. \\
    \graycell & Navigate to the receptacle that you picked the \{target\_object\_name\} from. \\    
    \graycell \multirow{-7}{*}{\textbf{\shortstack{Conditional\\Interaction}}}
    & \graycell Navigate to the object that you picked from the \{target\_receptacle\_name\}. \\
    
    & Navigate back to a \{target\_shape\} shaped object that you interacted with yesterday. \\
    & \graycell Navigate back to a \{target\_color\} colored object that you interacted with yesterday. \\
    & Navigate to an interacted object with \{target\_print\_or\_design\} on it. \\
    & \graycell Find an already interacted object that is made of \{target\_material\}. \\
    \multirow{-5}{*}{\textbf{Object Attributes}}
    & Go back to an interacted object that is used for \{target\_functionality\}. \\

    \graycell & \graycell Navigate to the receptacle that you interacted with which is the farthest from your current location. \\
    \graycell & Navigate to the receptacle that you did not interact with which is the farthest from your current location. \\
    \graycell & \graycell Navigate to the receptacle that you picked an object from which is the farthest from your current location. \\
    \graycell & Navigate to the receptacle that you placed an object on which is the farthest from your current location. \\    
    \graycell \multirow{-5}{*}{\textbf{Spatial Relationship}} & \graycell Navigate to the object which you interacted with which is the farthest from your current location. \\

    & Navigate to the room where you picked the \{target\_appearance\_order\} object from. \\
    & \graycell Navigate to the room where you placed the \{target\_appearance\_order\} object in. \\
    & Navigate to the room where you picked the \{target\_object\_name\} from. \\
    & \graycell Navigate to the room where you placed the \{target\_object\_name\} in. \\
    \multirow{-5}{*}{\textbf{Room Visitation}}
    & Navigate to a room that you did not visit yesterday. \\
    
    \graycell & \graycell Navigate to the \{target\_appearance\_order\} object that you interacted with yesterday. \\
    \graycell & Navigate to the \{target\_appearance\_order\} receptacle that you picked an object from. \\
    \graycell & \graycell Navigate to the \{target\_appearance\_order\} receptacle that you placed an object on. \\
    \graycell & Navigate to the receptacle that you picked the \{target\_appearance\_order\} object from. \\

    \graycell & \graycell Navigate to the object that you picked from the \{target\_appearance\_order\} receptacle. \\
    \graycell & Navigate to the object you interacted with immediately after ending the interaction with \{target\_object\_name\}. \\
    \graycell & \graycell Navigate to the object you interacted with immediately before interacting with \{target\_object\_name\}. \\
    \graycell & Navigate to the object you interacted with \{target\_num\_counts\} interactions after \{target\_object\_name\}. \\
    \graycell & \graycell Navigate to the object you interacted with \{target\_num\_counts\} interactions before \{target\_object\_name\}. \\
    \graycell & Navigate to the object that you interacted with between the interactions with \{target\_object\_name\_1\} and \{target\_object\_name\_2\}. \\

    \graycell & \graycell Navigate to the receptacle that you placed an object on right before you started interacting with \{target\_object\_name\}. \\
    \graycell & Navigate to the receptacle that you picked an object from right after you finished interacting with \{target\_object\_name\}. \\
    \graycell & \graycell Navigate to the receptacle that you placed an object on \{target\_num\_counts\} interactions before you started interacting with \{target\_object\_name\}. \\
    \graycell \multirow{-14}{*}{\textbf{Interaction Order}} & Navigate to the receptacle that you picked an object from \{target\_num\_counts\} interactions after you finished interacting with \{target\_object\_name\}. \\
    \graycell & \graycell Navigate to the receptacle that you placed an object on between the interactions with \{target\_object\_name\_1\} and \{target\_object\_name\_2\}. \\
    \graycell & Navigate to the receptacle that you picked an object from between the interactions with \{target\_object\_name\_1\} and \{target\_object\_name\_2\}. \\

    & \graycell Navigate to the receptacle that you interacted with at \{XX:XX\} yesterday. \\
    \multirow{-2}{*}{\textbf{Time-Based}}
    & Navigate to the object that you interacted with at \{XX:XX\} yesterday. \\

    \graycell & \graycell Navigate to the object which took the longest time to rearrange. \\
    \graycell & Navigate to the room that you spent the most time in. \\
    \graycell \multirow{-3}{*}{\textbf{Duration Tracking}}
    & \graycell Navigate to the object which took the shortest time to rearrange. \\

    & Revisit all the receptacles you picked objects from yesterday. \\
    & \graycell Revisit all the receptacles you placed objects on yesterday. \\
    & Revisit all the \{target\_receptacle\_name\} you placed objects on yesterday. \\
    & \graycell Revisit all the \{target\_receptacle\_name\} you picked objects from yesterday. \\
    & Revisit all the objects you interacted with yesterday. \\
    \multirow{-6}{*}{\textbf{\shortstack{Unordered\\Revisitation}}}
    & \graycell Revisit all the receptacles you interacted with yesterday. \\

    \graycell & Revisit all the receptacles you picked objects from yesterday in specific order. \\
    \graycell & \graycell Revisit all the receptacles you placed objects on yesterday in specific order. \\
    \graycell \multirow{-3}{*}{\textbf{\shortstack{Ordered\\Revisitation}}} 
    & Revisit all the objects you interacted with yesterday in specific order. \\
\end{longtable}

\end{document}